\documentclass{article}

\usepackage{arxiv}

\usepackage[utf8]{inputenc} 
\usepackage[T1]{fontenc}    
\usepackage{hyperref}       
\usepackage{url}            
\usepackage{booktabs}       
\usepackage{amsfonts}       
\usepackage{nicefrac}       
\usepackage{microtype}      
\usepackage{lipsum}
\usepackage{graphicx}
\usepackage{subcaption}
\graphicspath{ {./images/} }
\usepackage[acronym]{glossaries}
\usepackage{float}

\usepackage{algorithm}
\usepackage{algorithmic}

\newacronym{ml}{ML}{machine learning}
\newacronym{lstm}{LSTM}{Long Short Term Memory}
\newacronym{lstmae}{LSTM-AE}{LSTM Autoencoder}
\newacronym{rnn}{RNN}{Recurrent Neural Network}
\newacronym{ae}{AE}{Autoencoder}
\newacronym{bfast}{BFAST}{Breaks for Additive Season and Trend}
\newacronym{ewma}{EWMA}{exponentially weighted moving average}

\title{Early Detection of Forest Calamities in Homogeneous Stands -\\Deep Learning Applied to Bark-Beetle Outbreaks}

\author{
Maximilian Kirsch \\
  Intelligent System Research Group\\
  University of Applied Science Karlsruhe\\
   76187, Karlsruhe Germany \\
   \And
 Jakob Wernicke \\
 ThüringenForst AöR \\ 
 Forstliches Forschungs- \\ und Kompetenzzentrum (FFK) Gotha \\
 99867, Gotha Germany \\
  \And
 Pawan Datta \\
 ThüringenForst AöR \\ 
 Forstliches Forschungs- \\und Kompetenzzentrum (FFK) Gotha \\
 99867, Gotha Germany \\
 \And
 Christine Preisach \\
  Intelligent System Research Group\\
  University of Applied Science Karlsruhe\\
   76187, Karlsruhe Germany \\
}

\begin{document}
\maketitle
\begin{abstract}
Climate change has increased the vulnerability of forests to insect-related damage, resulting in widespread forest loss in Central Europe and highlighting the need for effective, continuous monitoring systems. Remote sensing based forest health monitoring, oftentimes, relies on supervised machine learning algorithms that require labeled training data. Monitoring temporal patterns through time series analysis offers a potential alternative for earlier detection of disturbance but requires substantial storage resources. This study investigates the potential of a Deep Learning algorithm based on a \gls{lstm} Autoencoder for the detection of anomalies in forest health (e.g. bark beetle outbreaks), utilizing Sentinel-2 time series data. This approach is an alternative to supervised machine learning methods, avoiding the necessity for labeled training data. Furthermore, it is more memory-efficient than other time series analysis approaches, as a robust model can be created using only a 26-week-long time series as input. In this study, we monitored pure stands of spruce in Thuringia, Germany, over a 7-year period from 2018 to the end of 2024. Our best model achieved a detection accuracy of 87\% on test data and was able to detect 61\%  of all anomalies at a very early stage (more than a month before visible signs of forest degradation). Compared to another widely used time series break detection algorithm - BFAST (Breaks For Additive Season and Trend), our approach consistently detected higher percentage of anomalies at an earlier stage. These findings suggest that LSTM-based Autoencoders could provide a promising, resource-efficient approach to forest health monitoring, enabling more timely responses to emerging threats.
\end{abstract}


\section{Introduction}
Forests in Central Europe are increasingly threatened by intensified insect outbreaks, largely driven by climate change. In Thuringia, Germany, for instance, approximately 20 \% of the forest area was affected by a bark beetle infestation during 2018 and 2023, eventually resulting in a timber loss of 28 Mio $m^3$ in the region \cite{thuringer_ministerium_fur_infrastruktur_und_landwirtschaft_waldzustandsbericht_2023}. That forest degradation in Central Germany but also in most of the spruce dominated forest regions in Central Europe is unprecedented. Climate change induces stressors that negatively impact the overall health and growth of forests \cite{randhir_emerging_2013, vacek_european_2023}. Particularly, hot summers and associated droughts impair the efficacy of defense mechanisms of spruce trees \cite{boisvenue_impacts_2006}.  Secondary pests like thermophilic bark beetle populations thrive under such circumstances, as their reproduction depends on predisposed and already weakened trees\cite{schwenke_beziehungen_1985, bentz_climate_2010}. \par

Predicting disturbances is challenging due to the unpredictable nature of insect induced calamities \cite{sergio_animal_2018}. An early and spatially precise detection of potential bark beetle infection center is needed to mitigate an epidemic bark beetle reproduction \cite{hlasny_bark_2021, kautz_early_2024}. Accordingly, several studies stressed the hypothesis of green attack detection based on remote sensing data \cite{abdullah_sentinel2_2019, barta_early_2021, barta_comparison_2022}. 
However, there is wide consensus that green attack detection is either unreliable or infeasible on large spatial scales \cite{lausch_forecasting_2013, wulder_challenges_2009,kautz_early_2024}. 
Instead, our focus is on identifying anomalous patterns in forest health that may serve as early warning indicators of potential disturbances. 
A limitation of early disturbance detection is often the lack of labeled training data. Satellite Image Time Series (SITS) based anomaly detection is a often unsupervised and does not rely on labeled training samples to capture seasonal variations \cite{verbesselt_detecting_2010, watts_effectiveness_2014}. Often SITS approaches, such as the BFAST algorithm, learn temporal patterns and detect deviations from the normal state \cite{verbesselt_bfast_2009}. Other approaches utilize deep learning such as Long Short-Term Memory neural networks, that remember temporal patterns in time series \cite{kong_long_2018}. 
Most approaches dependent on extensive historical information for every observed time series and often monitor univariate time series, creating hyperspectral relationships only by calculating vegetation indices (VIs) \cite{verbesselt_detecting_2010, watts_effectiveness_2014, kong_long_2018}. 

We investigated the potential of a deep learning, reconstruction-based time series anomaly detection to create an early warning system based on Sentinel-2 satellite images. We built a Long short-term memory (LSTM)-Autoencoder model for real-time anomaly detection with a 10-meter spatial resolution. We applied our approach to data collected from severely threatened spruce stands in Thuringia, Germany, over a period from 2018 to the end of 2024. Through observation and measurement of the health dynamics of the stands, we aim to flag anomalies in early disturbabce stages wall-to-wall. Our algorithm employs unsupervised deep learning and is independent from labeled training data. This makes our method accessible for a wide range of users, especially paired with freely available Sentinel-2 data. In addition, unlike many SITS approaches, this technique minimizes the need for extensive historical data, which is particularly important for single cell-based analyses in order to reduce memory demands.
The implementation of this early warning  system is potentially suitable for an operational forest disturbance monitoring and thus might support  forest health management strategies.\par

\textbf{Our main contributions are:} \par

\begin{enumerate}
\item{We propose a robust unsupervised approach for SITS anomaly detection in forest health dynamics that is highly accurate in detecting anomalies meaningfully prior to static methods based on vegetation index thresholds, by leveraging the temporal relationship between the multispectral channels on a single cell basis.}
\item{We introduce an approach for online anomaly detection, leveraging freely available Sentinel-2 satellite data. In contrast to conventional SITS methods, that require extended historical time series, our method processes incoming data in real time, mitigating the need for substantial temporal archives while maintaining robust detection performance.}
\item{We analyzed the importance of temporal information by testing LSTM-Autoencoder models that observe diverse window sizes from one month to one year.Thereby balancing model performance and historical data requirements to reduce storage demands.}
\item{We compared the \gls{lstm} models against a SITS method for break detection in time series\gls{bfast}, which was tested on diverse VIs. Our model achieved higher precision, accuracy and remarkable increase in early predictions.}
\item{Ultimately we propose a scoring method that allows us to compare diverse models' anomaly detection and early warning potential exceeding the generally used performance metrics, by rewarding earlier detections.}
\end{enumerate}


\section{Related Work}\label{Related Work}
Early detection of bark beetle outbreaks is a critical research area due to their significant impact on forest ecosystems. Detecting infested trees during the "green attack" stage - when beetles have infested a tree but crown symptoms are not yet visible - is particularly challenging, even in situ. This stage precedes the "red attack," where canopy discoloration becomes detectable in aerial imagery \cite{wulder_challenges_2009}.

\subsection{Remote Sensing-Based Bark Beetle Detection}

Classifying green attack using single remote sensing images is considered infeasible with medium-resolution satellite data \cite{wulder_challenges_2009}. Some studies have explored machine learning approaches to distinguish green attack from healthy vegetation. Abdullah et al. (2019) \cite{abdullah_sentinel2_2019} analyzed Sentinel-2 spectral bands and VIs, using statistical machine learning (ML) for variable selection. However, the results were deemed unsuitable for forestry applications by the authors \cite{abdullah_sentinel2_2019}. 
Other studies focussed on changes in forest stress levels using random forest classification, to distinguish bark beetle infested trees from healthy vegetation \cite{huo_early_2021}. These supervised machine learning approaches rely on labeled training data. The difficulty of creating reliable labels from satellite data, makes supervised machine learning impractical.

\subsection{Satellite Image Time Series Analysis}
Satellite image time series analysis (SITS) is widely used to detect forest disturbances, including bark beetle outbreaks \cite{barta_early_2021, abdullah_timing_2019, low_phenology_2020}. A key advantage of SITS is its ability to retrospectively analyze patterns leading to disturbances, such as red attack. SITS has been applied to classify bark beetle infestations using machine learning techniques. For instance, Barta et al. (2021) \cite{barta_early_2021} used a random forest model on SITS to pinpoint early separability between healthy and infested trees across various indices. They identified Sentinel-2 spectral bands and VIs that best differentiate healthy from infested trees. Abdullah et al. (2019) \cite{abdullah_timing_2019} explored the relationship between bark beetle and canopy chlorophyll content derived from SPOT-5 and RapidEye SITS data in combination with ground truth data. \par 
Anomaly detection methods are also prevalent in SITS research. An approach, introduced by Dutrieux et al. (2021) \cite{dutrieux_mise_2021} proposes a model, that detects time series harmonics and forecasts the pattern and if the actual time series deviates strongly from that pattern, an anomaly is flagged. The well known break detection model \gls{bfast}, detects changes by decomposing time series into trend, seasonal, and residual components\cite{verbesselt_bfast_2009}. This approach is commonly utilized to detect structural changes in vegetation time series \cite{xu_time_2022, watts_effectiveness_2014}. Furthermore, the sequential analysis method CUSUM has been employed to identify deviations in attacked trees from standard SITS patterns \cite{jamali_examining_2023, low_phenology_2020}. 

\subsection{Deep Learning Approaches}
Deep learning has gained traction in SITS for forest health classification and disturbance detection \cite{kong_long_2018, yuan_using_2020, zhou_autoencoder-based_2021}. Zhou et al. (2021) \cite{zhou_autoencoder-based_2021} proposed an unsupervised \gls{ae} model that detects anomalies in univariate Landsat time series by identifying reconstruction errors. \gls{rnn}, particularly using \gls{lstm} architectures, excel at capturing temporal dynamics in vegetation phenology and outperform traditional methods in land cover classification \cite{ruswurm_multi-temporal_2017, ruswurm_self-attention_2020}. LSTM models have been applied to change detection on Sentinel-1 data \cite{saha_change_2022} and burned area mapping using MODIS data \cite{yuan_using_2020, kong_long_2018}. Kong et al. (2018) \cite{kong_long_2018} introduced a LSTM-based framework that forecasts time series and flags disturbances when predictions deviate significantly from observations. Recent research highlights the potential of multivariate \gls{lstmae} for anomaly detection in other fields, such as cybersecurity \cite{wei_reconstruction-based_2023} or water quality monitoring across multiple features (variables) \cite{li_anomaly_2024}.

Our work investigates the applicability of multivariate \gls{lstmae} models on Sentinel-2 SITS for spruce forest monitoring. We focus on identifying broader anomalies in forest health through remote sensing-based anomaly detection. By leveraging multivariate temporal relationships, we aim to detect anomalies in the temporal and multispectral patterns to predict disturbances while minimizing false positives. Additionally, we evaluate various input window sizes to balance model performance and historical data requirements, thereby reducing storage and computing demands. 
In contrast, existing work predominantly focuses on univariate anomaly detection models requiring extensive historical data. Limited attention has been given to multivariate approaches using reconstruction-based \gls{lstmae} methods on Sentinel-2 data. 
Our study seeks to address this gap by enabling early disturbance detection with reduced false positives and optimized data storage demands.

\section{Study Area and Data}
In this chapter, we describe our two study areas, the larger study area, the Thuringian Forest and the Südharz Nature Park, which are the selected areas for training and testing, and the Sentinel-2 time series data basis.


\begin{figure}[H]
  \centering
  \begin{subfigure}[b]{0.5\textwidth}
    \centering
    \includegraphics[width=\linewidth]{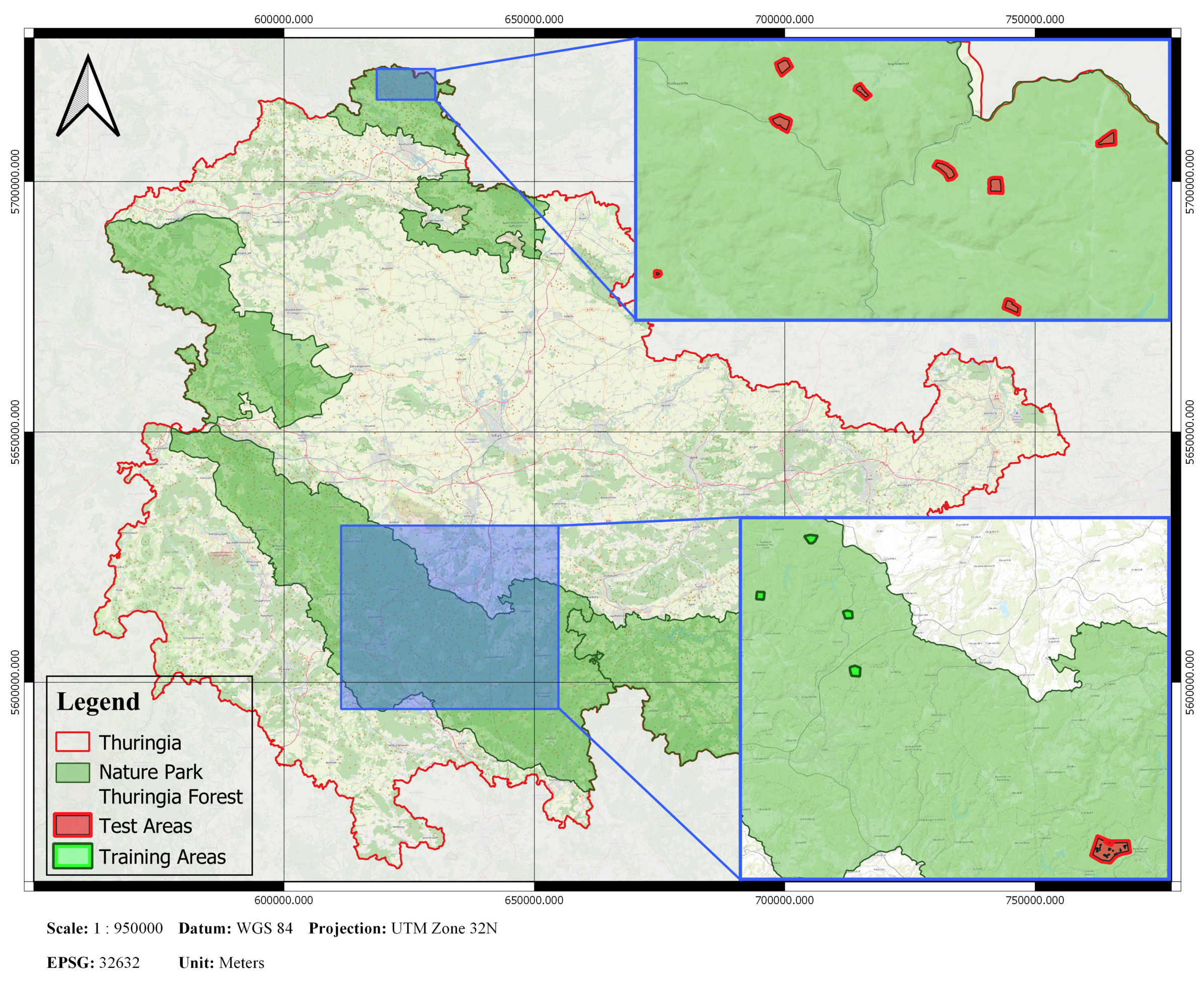}
    \caption{Locations of the Study Area in Thuringia, Germany.}
    \label{th_map}
  \end{subfigure}
  \hfill
  \begin{subfigure}[b]{0.45\textwidth}
    \centering
    \includegraphics[width=\linewidth]{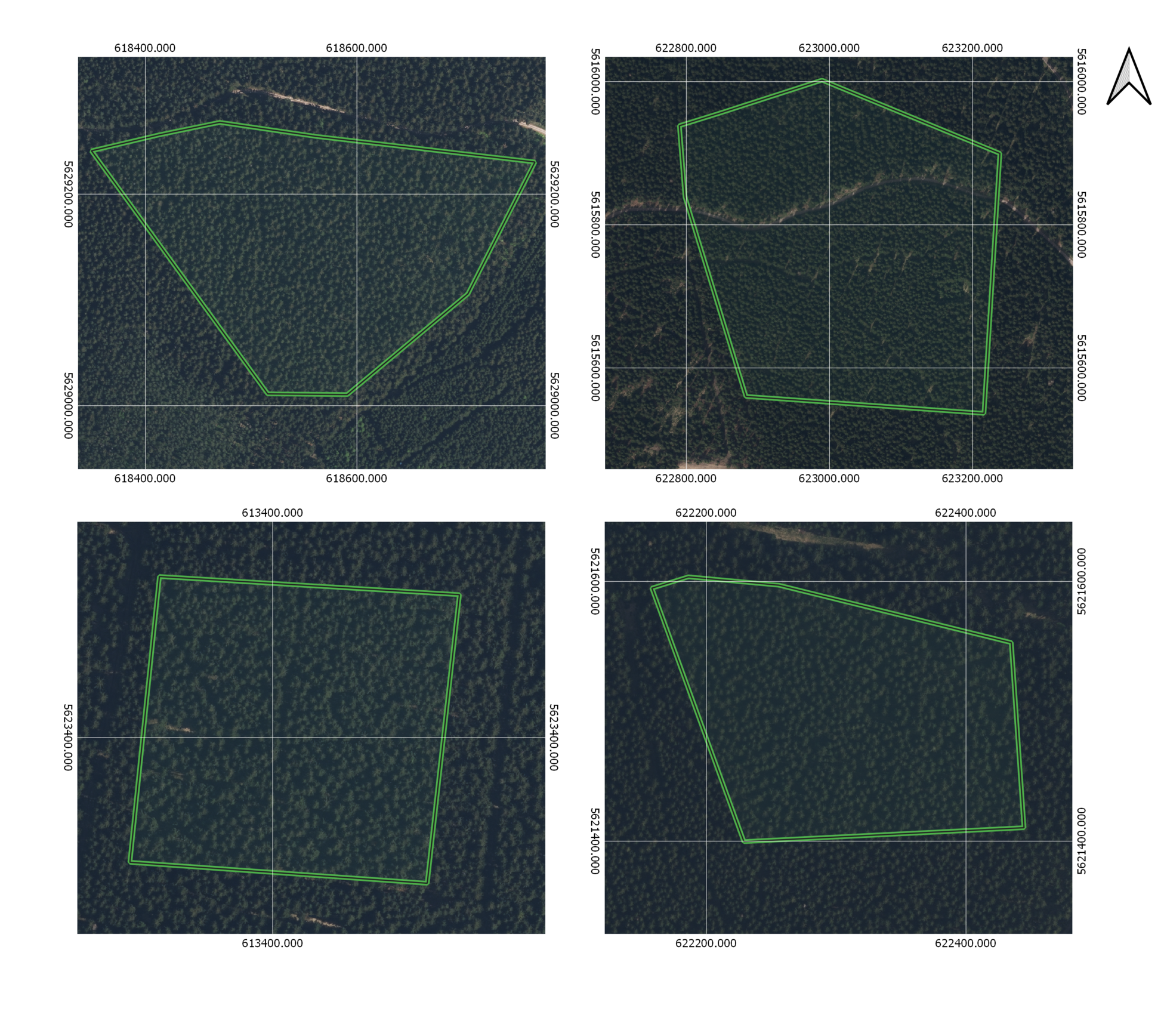}
    \caption{Areas Selected as Training Data. The images show recordings from June 4, 2023.}
    \label{training_area_image}
  \end{subfigure}
  \caption{Overview of the study area and training data.}
  \label{fig:combined}
\end{figure}

\subsection{Study Area} \label{subs_studyArea}
To evaluate the potential of LSTM Autoencoder for early disturbance detection, we retrospectively observed spruce stands in the Thuringian Forest and the southern part of the Harz mountains (central Germany) during the period January 2018 to December 2024. Thuringia is a suitable study area for the detection of disturbances due to its extensive bark-beetle induced forest degradation since 2018. The predominately managed forests of Thuringia are characterized by three leading tree species (spruce, beech, pine), which were partly intermixed with secondary tree species. Before 2018 even-aged singled layer spruce stands exhibited the typical forest structure in the region. As mentioned previously, the persistent drought since 2018 initiated the most severe forest degradation in the region. As a result, approximately 110.000 ha of mainly spruce stands were clear cut due to the mass outbreak of bark beetle \cite{thuringer_ministerium_fur_infrastruktur_und_landwirtschaft_waldzustandsbericht_2023}. \par

We used true color orthorectified aerial images (orthophotos at a spatial resolution of 0.2 meters) published by the state of Thuringia \footnote{URL to the orthophotos: https://geoportal.thueringen.de/gdi-th/download-offene-geodaten/
download-luftbilder-und-orthophotos. Last accessed on 30 August 2024} recorded on June 4, 2023, to visually analyze and select areas corresponding to pure spruce stands with varying degrees of health.  

We selected four areas for training as well as a larger area for testing in the Thuringia Forest region. Additionally, eight test areas were chosen from the the Südharz Nature Park, located in the northern part of Thuringia (Figure:  \ref{th_map}).

The model was trained using time series of healthy spruce stands. For this, we selected visually healthy forest patches with green crowns as training areas, covering a total of 28.85 ha. The health status of these areas throughout the observation period was confirmed by analyzing the time series of VIs (NDVI, DWSI). \par The selected training areas are shown in Fig. \ref{training_area_image}.


The areas designated for testing depict varying degrees of disturbance (e.g., green forest, red crowns, and clear-cuts). The polygonal areas varied in size from 25 to 561 pixels. Furthermore, we selected a larger test region in which various stages of disturbance were represented in a contiguous area of 44,519 pixels (455 ha). We used this region to create an anomaly map.

\subsection{Sentinel‐2 Time Series} \label{ssec:data_ts}
The Sentinel‐2 data were selected for their high spatial resolution, frequent revisit time, and free availability. These data capture high-resolution images of the Earth’s surface with a combined revisit interval of five days \cite{nakar_sentinel-2_2019}.

We selected atmospherically corrected Sentinel-2 Level 2A (L2A) data, which also include a scene classification map \cite{louis_level-2a_2021}. The Scene Classification (SCL) algorithm characterizes every pixel in a Sentinel‐2 image into one of 12 classes \cite{campos-taberner_understanding_2020}. These classes describe cloud types, snow, vegetation, or non-vegetated areas, which we utilized to filter instances where an observed pixel was covered by cloud or snow at a given time step. \par
The Sentinel‐2 L2A data is represented by multivariate time series for each pixel, covering 12 spectral bands from January 2018 to December 2023. These were organized into spatially and temporally indexed tables.

\section{Background: LSTM Autoencoder Model}
In this section, we provide a comprehensive overview of the model components and the model architecture of the unsupervised Deep Learning algorithm we used for early detection of bark-beetle outbreaks. First, the general functionality of \gls{ae}'s is set out, followed by the operating principles of LSTM. In conclusion, we outline the general functionality and structure of the LSTM \gls{ae} model.  
\subsection{Autoencoder}
We built an \gls{ae} for anomaly detection. An \gls{ae} is a neural network designed to reconstruct input data using two functions: an encoder and a decoder \cite{wei_reconstruction-based_2023}. The encoder $E$ learns the most important features of the input data $x$ and reduces them into a lower-dimensional representation called the latent space $z$ \cite{li_comprehensive_2023}:
\begin{equation}
    z = E(x)
\end{equation}

\begin{figure}[H]
    \centering
    \includegraphics[width=0.6\linewidth]{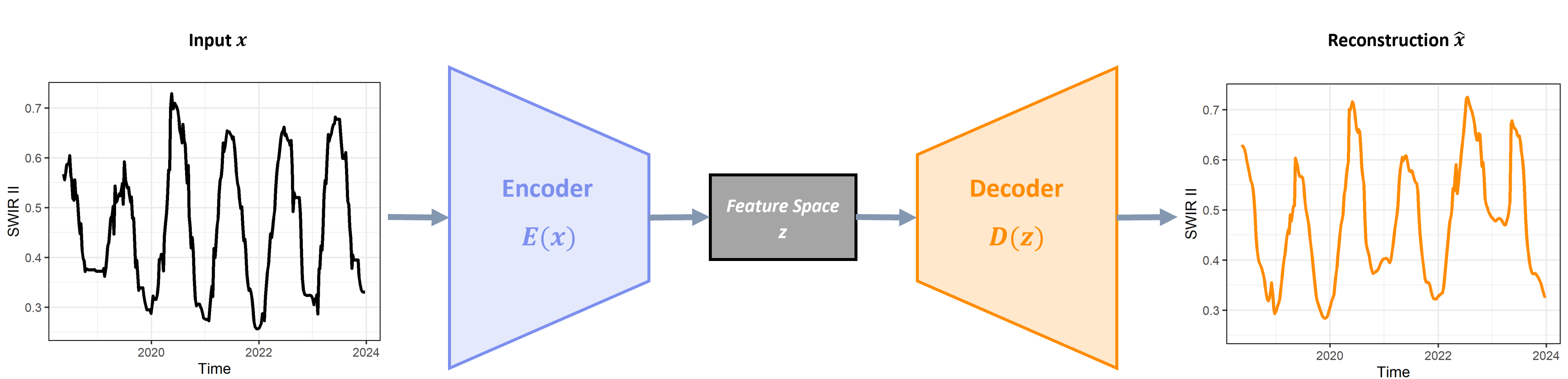}
    \hfil
    \caption{Autoencoder: Reconstruction of a time series.}
    \label{fig_ae}
\end{figure}

The decoder $D$ converts the latent space representation $z$ back to the original shape, producing the reconstructed data $\hat{x}$ \cite{li_comprehensive_2023}:
\begin{equation}
    \hat{x} = D(z)
\end{equation}
The structure of an \gls{ae} and the process of reconstruction is illustrated in Fig. \ref{fig_ae}.

During training, the AE optimizes parameters, such as weights and biases, to minimize the reconstruction error between the input $x$ and the reconstruction $\hat{x}$ \cite{goodfellow_deep_2016}.
In anomaly detection, \gls{ae}'s are trained exclusively on 'normal' data samples, without anomalies and learn to capture the key features \cite{ma_deep_2019, trinh_detecting_2019}. The difference between input data $x$ and 
reconstructed data $\hat{x}$ is known as the reconstruction error \cite{goodfellow_deep_2016}. Since anomalies exhibit atypical characteristics, they cannot be accurately reconstructed, leading to high reconstruction errors \cite{chandola_anomaly_2009}. Anomalies are identified when the reconstruction error exceeds a predefined threshold \cite{chandola_anomaly_2009}.

\subsection{LSTM} \label{subs_lstm}
\begin{figure}[H]
    \centering
    \includegraphics[width=0.6\linewidth]{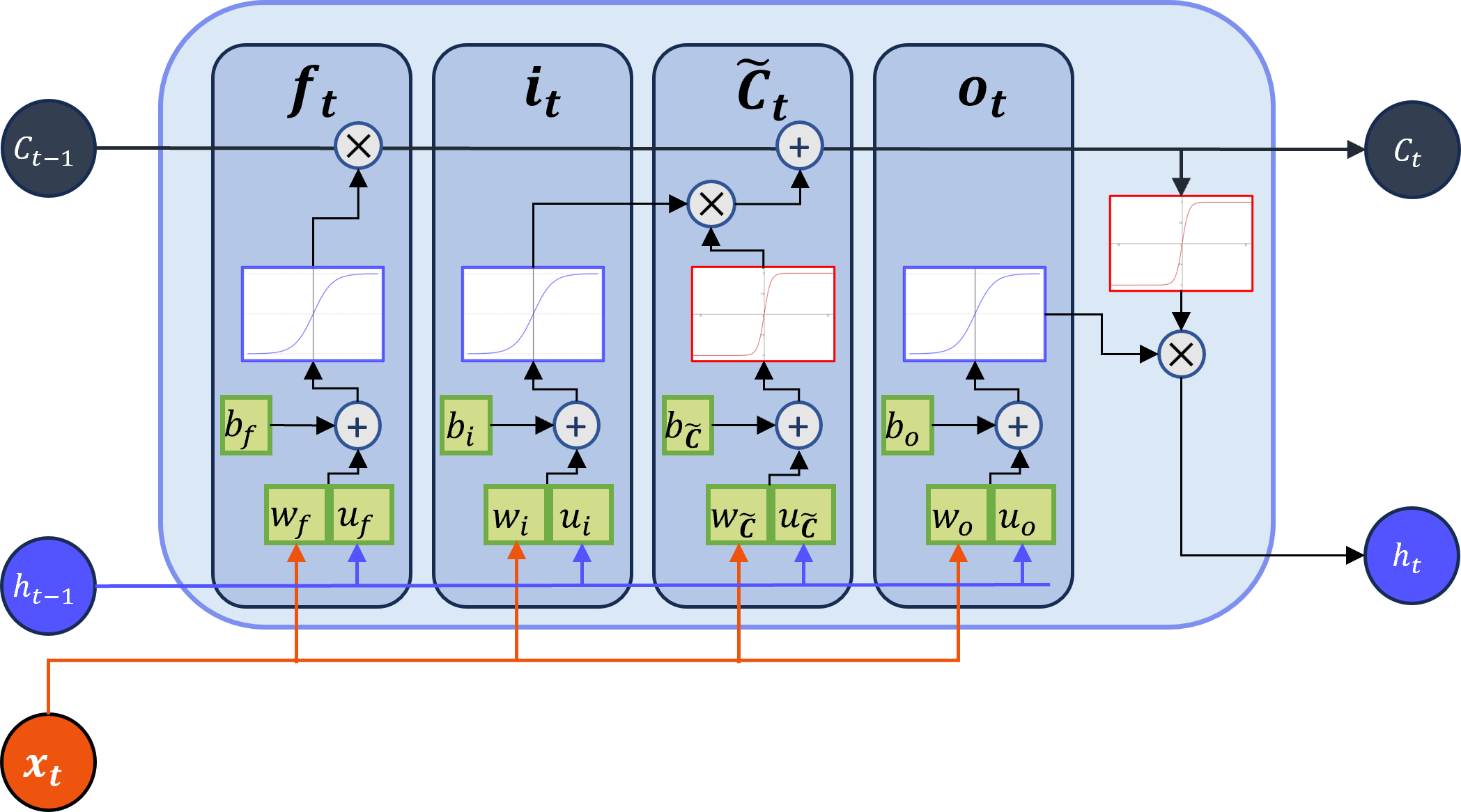}
    \hfil
    \caption{Components of a LSTM-Unit.}
    \label{lstm_structure}
\end{figure}

We incorporated Long Short-Term Memory (LSTM) into our Autoencoder model, making it an effective method for reconstruction-based anomaly detection in time series \cite{wei_reconstruction-based_2023}. LSTM is a type of Recurrent Neural Network (\gls{rnn}), that is designed to efficiently process sequential data e.g. time series. This ensures that temporal relationships are preserved during processing \cite{wei_reconstruction-based_2023}. The LSTM architecture includes a component for maintaining long-term dependencies and another for propagating short-term memories \cite{russell_artificial_2020}. At each time step, new information is added, updating the memory of the cell \cite{russell_artificial_2020}.
The LSTM unit architecture that we applied is an adaptation of the original LSTM model from \cite{hochreiter_long_1997} and incorporates a mechanism known as the forget gate. This advanced LSTM cell, proposed by \cite{gers_learning_2000}, adds a weighted function that determines which information from the input should be discarded. The LSTM architecture is illustrated in Fig. \ref{lstm_structure}.

Figure \ref{lstm_structure} depicts the components within a basic LSTM cell, as well as the structural flow of the data. The Cell State, denoted \( C \), represents the long-term memory, while the Hidden State, denoted \( h \), stores information about the short-term memory. The input data \( x_t \) describes the values for the multivariate time series at the current time step. Furthermore, the LSTM unit contains so called gated units, these are the forget gate \( f_t \), input gate \( i_t \), cell candidate gate \( \tilde{C}_t \) and output gate \( o_t \). The weights \( w \), \( u \), and the bias \( b \) are learned on the input data and the hidden state, and an activation function \( \sigma \) or \( \tanh \) is applied.
\( \sigma \) is displayed with a blue frame and \( \tanh \) red.
\bigskip

Each LSTM unit has three inputs: (1) The input data of the current time step \(x_t\), (2) the Cell State of the previous time step \(C_{t-1}\), and (3) the Hidden State of the previous time step \(h_{t-1}\). Furthermore, it creates a new Cell State \(C_t\) and a new Hidden State \(h_t\) as outputs. 

The gated units, as well as the information forward pass of an LSTM unit, are defined by the following equations: 

\begin{equation}
    f_t = \sigma\left(w_f x_t + u_f h_{t-1} + b_f\right) 
    \label{eq:forget_gate}
\end{equation}
\begin{equation}
    i_t = \sigma\left(w_i x_t + u_i h_{t-1} + b_i\right) 
    \label{eq:input_gate}
\end{equation}
\begin{equation}
    o_t = \sigma\left(w_o x_t + u_o h_{t-1} + b_o\right) 
    \label{eq:output_gate}
\end{equation}
\begin{equation}
    \tilde{C}_t = tanh \left(w_{\tilde{C}} x_t + u_{\tilde{C}} h_{t-1} + b_{\tilde{C}}\right) 
    \label{eq:ig_2}
\end{equation}
\begin{equation}
    C_t = f_t \odot C_{t-1}+i_t \odot \tilde{C}_t 
    \label{eq:updated_LTM}
\end{equation}
\begin{equation}
    h_t = o_t\odot tanh\left(C_t\right) 
    \label{eq:ht_out}
\end{equation}

Where \(\odot\) denotes the point-wise multiplication of two vectors. 

The sigmoid activation function is calculated as: 

\begin{equation}
    \sigma(x) = \frac{1}{1 + e^{-x}}
    \label{eq:sigmoid}
\end{equation}
\par
The tanh activation function is calculated as:
\begin{equation}
    tanh(x) =  \frac{e^x - e^{-x}}{e^x + e^{-x}}
    \label{eq:tanh}
\end{equation}


The operational mechanics of this LSTM unit, as detailed by \cite{yu_review_2019}, are as follows: 

The equations \ref{eq:forget_gate}, \ref{eq:input_gate}, and \ref{eq:output_gate} calculate the gated units. Each gated unit vector multiplies the input at the current time step \(x_t\) with the respective weights \(w\) and multiplies the Hidden State of the previous time step \(h_{t-1}\) with the respective recurrent weight \(u\). The vectors are then summed before a bias is added. The results are then passed through a sigmoid activation function, producing a value between 0 and 1. The gated units differ only in terms of their respective weights and biases, allowing the LSTM cell to forget irrelevant information of a sequence \cite{goodfellow_deep_2016, russell_artificial_2020}.
\par\medskip

\par
Equation \ref{eq:ig_2}, the \emph{cell candidate gate} defines the potential new Cell State \(\tilde{C}_t\). Likewise to the equation of the gated units, \(x_t\) and \(h_{t-1}\) are multiplied with the respective weights \(w_C\) and \(u_C\). The products are summed and a bias \(b_C\) is then added. The result serves as input for a tanh activation function
that regulates the information flow in the network.
\par
To determine the new Cell State \(C_t\), Equation \ref{eq:updated_LTM} takes the previous Cell State \(C_{t-1}\) and multiplies it by the output of the forget gate. Since \(f_t\) is a value between 0 and 1, it dictates the proportion of \(C_{t-1}\) that is retained. Subsequently, the potential new Cell State \(\tilde{C}_t\) is point-wise multiplied by \(i_t\) to determine the proportion of \(\tilde{C}_t\) that is added.
\par
Equation \ref{eq:ht_out} defines the new Hidden State \(h_t\). It is calculated by using the new Cell State \(C_t\) as input for a Tanh Function resulting in a potential new Hidden State. This is then point-wise multiplied by the output gate \(o_t\) to determine the proportion of potential Hidden State that is passed on to the next time step.
\par

\subsection{LSTM-Autoencoder Architecture}

\begin{figure}[H]
    \centering
    \includegraphics[width=0.9\linewidth]{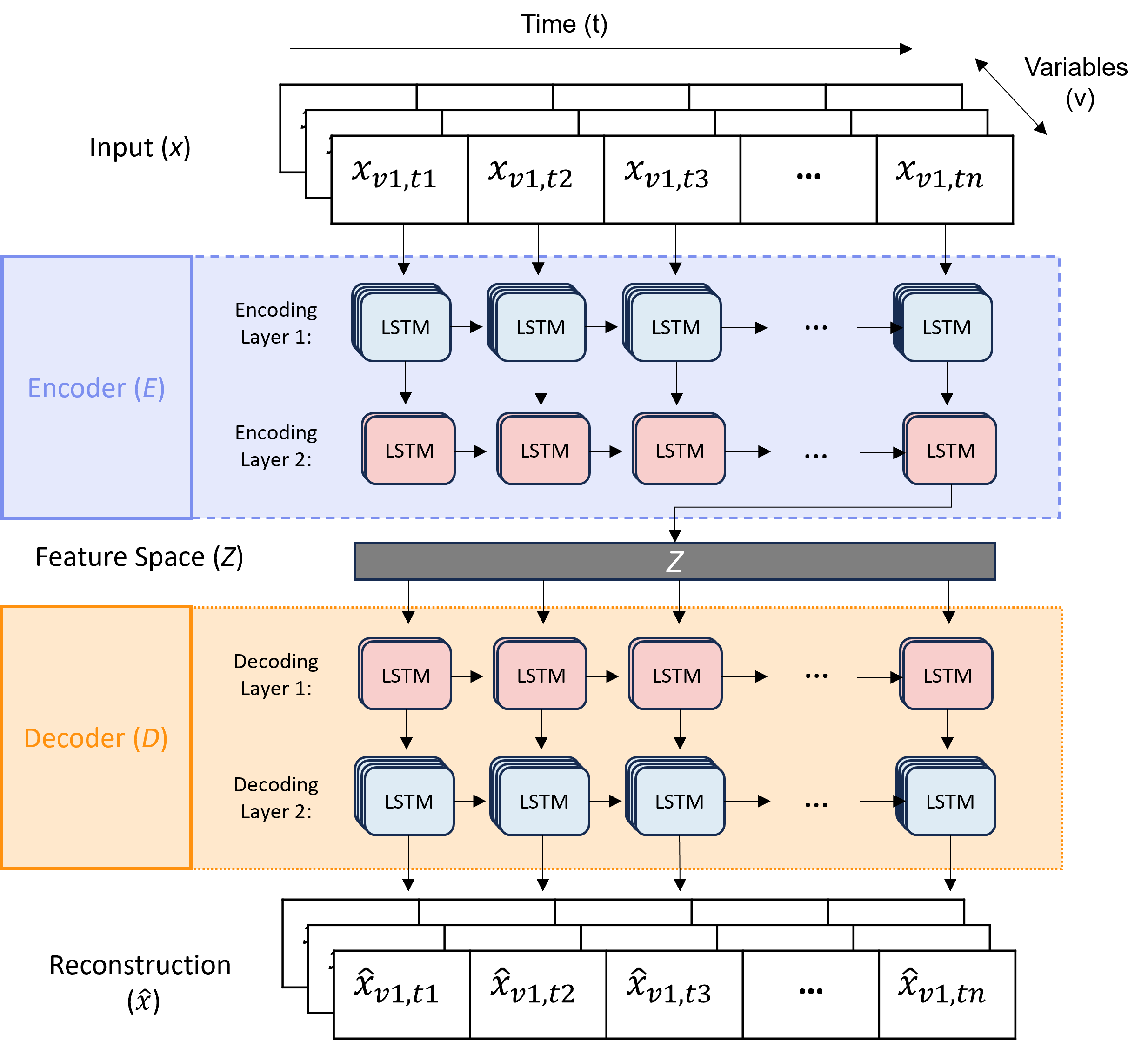}
    \hfil
    \caption{LSTM Autoencoder Architecture.}
    \label{lstm_ae}
\end{figure}

We incorporate stacked LSTM neurons into an \gls{ae} network (see Fig. \ref{lstm_ae}). LSTM units detect long- and short-term dependencies, as well as relationships between input features, making them highly effective for multivariate time series anomaly detection \cite{li_comprehensive_2023, wei_reconstruction-based_2023}.

We applied a Long Short-Term Memory Autoencoder (LSTM AE) architecture based on the method proposed by \cite{wei_reconstruction-based_2023}. They built an LSTM AE to detect DDoS attacks\footnote{DDoS stands for Distributed Denial of Service. DDoS attacks are cyberattacks aimed at overwhelming a server by sending enormous amounts of data \cite{wei_reconstruction-based_2023}.} in multivariate time series data.

Before applying the model to a time series, the data are divided into sequences of a predefined window size, as explained in \ref{preprocessing}. Within each sequence, the relationships between features and time steps are computed using hidden LSTM units \cite{wei_reconstruction-based_2023}.

Multiple LSTM units are stacked in each layer of the Autoencoder. Within each unit, the long-term memory $C_{t,i}$ (where $t$ represents time and $i$ denotes the unit index) and the short-term memory $h_{t,i}$ are continuously updated with every additional time step \cite{goodfellow_deep_2016}. The outputs $C_{t,i}$ and $h_{t,i}$ from each unit are then passed to all units in the subsequent Autoencoder layer. Since the second layer contains fewer LSTM units, the information is progressively compressed, reducing the dimensionality while retaining the most relevant features.  

The final LSTM layer in the encoding phase produces highly compressed representations of the multivariate input data \cite{goodfellow_deep_2016}, forming the latent space, which is denoted as $Z$. Before the data is passed to the decoder network, a \emph{Repeat Vector} layer is applied to reshape $Z$ to match the length of the input sequence. These repeated vectors then serve as inputs for the decoder network \cite{wei_reconstruction-based_2023}.  

During the decoding phase, the compressed information is gradually decompressed through successive layers, ultimately restoring the data to its original structure. A dense layer is applied to reshape the reconstructed output to match the original input dimensions \cite{wei_reconstruction-based_2023}. Throughout the training process, the model optimizes its weights and biases to minimize reconstruction error and accurately reconstruct the input multivariate time series sequence \cite{wei_reconstruction-based_2023, russell_artificial_2021}.  

The processing and data flow within an LSTM AE is illustrated in Fig. \ref{lstm_ae}.

Fig. \ref{lstm_ae} shows the way the LSTM AE encodes the multivariate input time series $x$ into a compressed representation of the input data, a latent layer, called feature space $Z$. The feature space is then decoded back to the input shape. The reconstruction is denoted as $\hat{x}$  .

\section{Early Detection Framework}
\begin{figure}[H]
    \centering
    \includegraphics[width=1\linewidth]{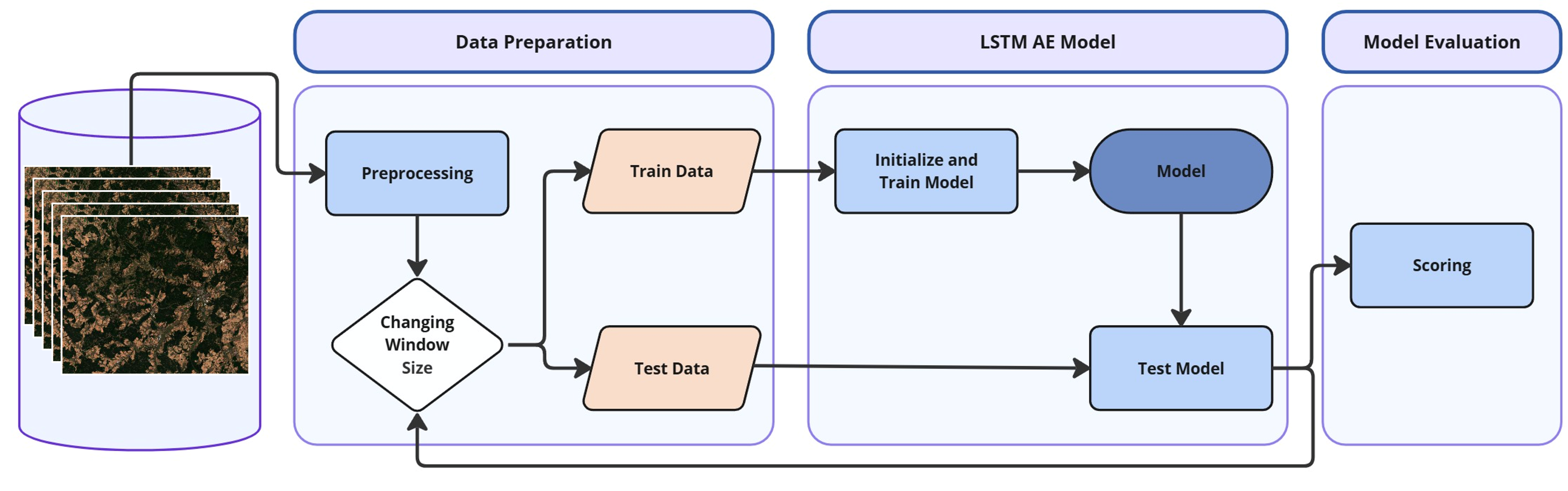}
    \hfil
    \caption{Methodology Outline}
    \label{methodologyFC}
\end{figure}

 In this section, we outline our proposed framework for early disturbance detection. We begin with data preparation, followed by the initialization, training and testing  of our LSTM AE model. Finally, we introduce a new evaluation framework. Fig. \ref{methodologyFC} shows the steps of the process we followed.
 
\subsection{Data Preparation} \label{preprocessing}
As mentioned in \ref{ssec:data_ts}, the data for all observed sites are contained in spatial and temporal indexed data frames. During preprocessing, the multivariate time series for each pixel was treated individually. 
\par
\smallskip
The time series contain a signal for every fifth day. Due to cloud or snow coverage, many data instances are deemed unsuitable and are excluded from the analysis. To filter these instances, we used the SCL layer from the L2A images to retain only pixels classified as either vegetated or non-vegetated. As shown in Equation \ref{decomposition_ts} we decomposed the time series into trend, seasonal component and remainder and applied Tukey's method \cite{tukey_exploratory_1977} to the remainder $R_t$ to remove remaining outliers. 

\begin{equation} \label{decomposition_ts}
   x_t = T_t + S_t + R_t
\end{equation}
Where $T_t$ denotes trend, $S_t$, seasonal component and $R_t$ the remainder.
We then aggregated the data to weekly values and imputed the gaps to obtain homogeneous time series. First, we smoothed the time series using the Savitzky-Golay filter (SGF) and then  filled the gaps with \gls{ewma}. The SGF smooths and differentiates data by performing a local least squares polynomial fit within a sliding window. It detects the underlying trend of the data while reducing noise. Applying SGF to enhance the quality of satellite image-based time series by smoothing out fluctuations and noise has been widely referenced \cite{cao_simple_2018, chen_simple_2004, feng_reconstruction_2023, kordestani_direct_2020}. A window size is selected, determining the number of observations used in the smoothing process. If the window consists of $2S+1$ points, the consecutive observations within the window are $x_{i - S}, x_{i - S+1}, ..., x_i, ..., x_{i+S}$. To model the trend within each window, a polynomial of order $r$ is fitted using least squares regression. The fitted polynomial function:

\begin{equation}
    F(k) = \sum_{j=0}^{r} \beta_j k^j 
\end{equation}

where $k$ denotes the relative position of a data point within the local window, centered at $i$, $\beta_j$ denotes the coefficients of the polynomial function and $r$ denotes their order.
The coefficients of  the polynomial function are then calculated using the least squares method by minimizing the following expression:

\begin{equation}
    min\left\{ \sum_{k=-S}^{S} \left( y_{i+k} - F(k) \right)^2 \right\} 
\end{equation}
The SGF uses all observations, so $x_{i-S}, x_{i-S+1}, ..., x_i, ..., x_{i+S}$ to determine a new value for $x_t$ at $t = i$. 

After smoothing the time series, the gaps were filled using \gls{ewma}-imputation, which fills missing values of a time series by calculating the mean from an equal number of observations on either side of a missing value. \gls{ewma} employs a weighting factor that decreases exponentially with the number of time steps from the missing value. The weighting factor is given by $w_j = 2^{-j}$, where $j$ represents the temporal distance from the missing value \cite{moritz_imputets_2017}.
\par
\smallskip
The LSTM AE model is designed to process sequences of multivariate time series. The length of the input sequences is referred to as the 'window size'. This defines the observation window of the model and thus the minimum amount of historical data required for the model to operate. For the initial run we selected a window size of 52 instances (one year) to cover an entire seasonal period. Both, healthy training time series and the testing time series, were preprocessed using the above mentioned methods and the same window size was selected.  

\subsection{LSTM AE Model-Training and Inference}
Here, we outline the development, training, and evaluation of our model. To assess performance across different temporal contexts, the procedure was carried out for multiple window sizes of 4, 12, 26, and 52 weeks. The complete process is detailed in Algorithm \ref{alg:anomaly_detection}.

\begin{algorithm}[H]
\caption{LSTM Autoencoder Anomaly Detection}
\label{alg:anomaly_detection}

\begin{algorithmic}
\STATE $Tr \gets$ Training dataset of time series of healthy coordinates
\STATE $Te \gets$ Test dataset of time series of healthy and unhealthy pixels 
\STATE $\tau \gets$ Anomaly detection threshold 
\STATE $ws \gets$ Window size 
\STATE $x \gets$ time series in $Tr$ or $Te$
\STATE $t \gets$ time step in $x$

\STATE \textbf{Train LSTM-AE:}  
\FOR{$ws \in \{4, 12, 26, 52\}$}  
    \FOR{$x \in Tr$}  
        \STATE $LSTM\_AE_{ws} \gets train\_model(x)$  
    \ENDFOR  
\ENDFOR  

\STATE \textbf{LSTM-AE inference:}  
\FOR{$ws \in \{4, 12, 26, 52\}$}  
    \FOR{$x \in Te$}  
        \STATE $\hat{x} \gets LSTM\_AE_{ws}(x)$    
        \begin{align*}
        Anomaly_{x,t} = 
        \begin{cases} 
            1, & \text{if } |x_{t} - \hat{x}_{t}| > \tau, \\
            0, & \text{otherwise.}
        \end{cases}
        \end{align*}
    \ENDFOR
\ENDFOR  
\end{algorithmic}
\end{algorithm}

\smallskip
\paragraph{Initialization} To initialize the model, we built a multilayer Autoencoder architecture that processes multivariate time series sequences.
We chose to select the following Sentinel-2 bands as features: B2, B3, B4, B5, B6, B7, B8, B11, B12, therefore the number of features is 9. The bands, their wavelengths, as well as their resolutions are listed in \emph{Table \ref{tab:bands}}. 

\begin{table}[H]
    \centering
    \caption{Sentinel-2 Bands, Description, Wavelength Range, and Resolution.}

        \resizebox{\linewidth}{!}{ 
    \begin{tabular}{c c c c}
    \hline
    \textbf{Band Number} & \textbf{Band Description} & \textbf{Wavelength Range (nm)} & \textbf{Resolution (m)} \\ \hline
    B2 & Blue & 458--523 & 10 \\ 
    B3 & Green & 543--578 & 10 \\ 
    B4 & Red & 650--680 & 10 \\ 
    B5 & Red-edge 1 & 698--713 & 20 \\ 
    B6 & Red-edge 2 & 733--748 & 20 \\ 
    B7 & Red-edge 3 & 773--793 & 20 \\ 
    B8 & NIR & 785--900 & 10 \\ 
    B11 & SWIR I & 1565--1655 & 20 \\ 
    B12 & SWIR II & 2100--2280 & 20 \\ 
    \hline
    \multicolumn{4}{l}{\footnotesize NIR = Near infrared; SWIR = Shortwave infrared} \\
    \end{tabular}
    }
    \label{tab:bands}
\end{table}

Our model architecture consists of three encoding layers with 256, 128, and 64 LSTM units per layer. A dropout layer with a dropout rate of 0.2 was inserted between the first and second LSTM layer to prevent overfitting of the model \cite{srivastava_dropout_2014}. First, a repeat layer was used in the decoder, which repeats the latent space vector to the initial window size. This is followed by two LSTM layers with 64 units and 128 units, as well as a dropout layer with a dropout rate of 0.2. Finally, another LSTM layer with 256 units is inserted into the architecture and a time distributed dense function layer, which compresses the decoder output back to the input size, is used for each time step individually\footnote{The model was defined using the open-source machine learning platform TensorFlow.}. The multilayer model structure with the number of trainable parameters in each layer is described in Table \ref{tab:Model1}.

\medskip
\begin{table}[H]
    \caption{LSTM-AE Model Structure}
    \centering
    \footnotesize
    \begin{tabular}{|l c r|}
    \hline
    \textbf{Layer} & \textbf{Output Shape} & \textbf{Parameters} \\ \hline 
    Model Input&     $\left(window\_size, 9\right)$ &  \\     \hline 
    \textbf{Encoder} & & \\
    \textbf{256} LSTM Units& $\left(window\_size, 256\right)$  &  272,384 \\
    Dropout Layer & $\left(window\_size, 256\right)$  & 0 \\
    \textbf{128} LSTM Units& $\left(window\_size, 128\right)$    & 197,120  \\
    \textbf{64} LSTM Units& $\left(64\right)$         & 49,408  \\ \hline
    Repeat Vector & $\left(window\_size, 64 \right)$    & 0 \\    \hline
    \textbf{Decoder} & & \\
    \textbf{64} LSTM Units& $\left(window\_size, 64 \right)$  & 33,024   \\
    \textbf{128} LSTM Units& $\left(window\_size, 128\right)$  & 98,816  \\
    Dropout Layer & $\left(window\_size, 128\right)$  & 0 \\
    \textbf{256} LSTM Units& $\left(window\_size, 256\right)$  & 394,240  \\
    Time Distribution & $\left(window\_size, 9\right)$  & 2,313  \\   \hline
    \end{tabular} 
    \label{tab:Model1}
\end{table}

\paragraph{Model Training}

\begin{figure}[H]
    \centering
    \includegraphics[width=0.6\linewidth]{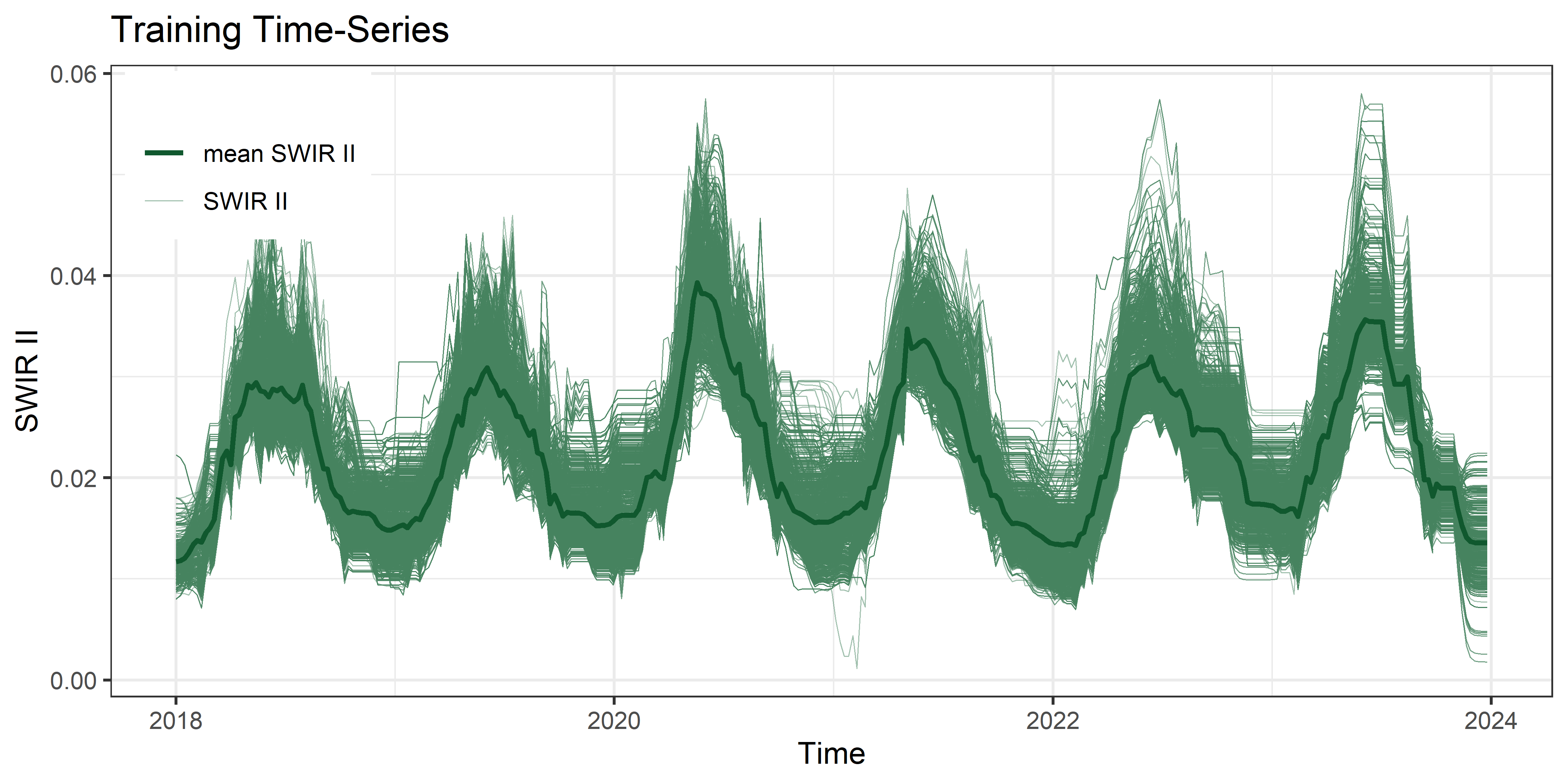}
    \caption{Short Wave Infrared II Training Data Samples}
    \label{test_dataset}
\end{figure}

After defining the model, the preprocessed time series of the healthy training areas, which were described  in \ref{subs_studyArea}, were visually analyzed, and their health status was assessed using VIs. Pixels exhibiting disturbances at any point were excluded, resulting in 2,885 time series from 2018 to 2024. Fig. \ref{test_dataset} shows one variable (band) of the training data.

Then, each variable is scaled between 0 and 1 and the time series is segmented according to a selected window size, initially set to 52 to encapsulate a full seasonal cycle. This results in 17,310 batches, each containing 52 time steps and 9 features.

Since the model operates in an unsupervised setting, no labels are required during training. The objective was to learn typical seasonal reflectance patterns of healthy spruce forests, enabling it to accurately reconstruct time series of healthy forests while recognizing anomalous patterns,such as bark-beetle infestations \cite{goodfellow_deep_2016, wei_reconstruction-based_2023}.

Each \gls{lstm} unit is defined by four input weights ($w_f, w_i, w_C, w_o$), four recurrent weights ($u_f, u_i, u_C, u_o$), and four bias values ($b_f, b_i, b_C, b_o$) \cite{gers_learning_2000}. The number of trainable parameters in an LSTM layer is determined by:
\begin{equation}
\theta = 4 \left( h(h + i) + h \right)
\end{equation}
The number of units in a layer is denoted as $h$ and the input features are denoted as $i$. $\theta$ represents the tunable parameters \cite{karim_counting_2019}. 

\begin{figure}[H]
    \centering
    \includegraphics[width=0.9\linewidth]{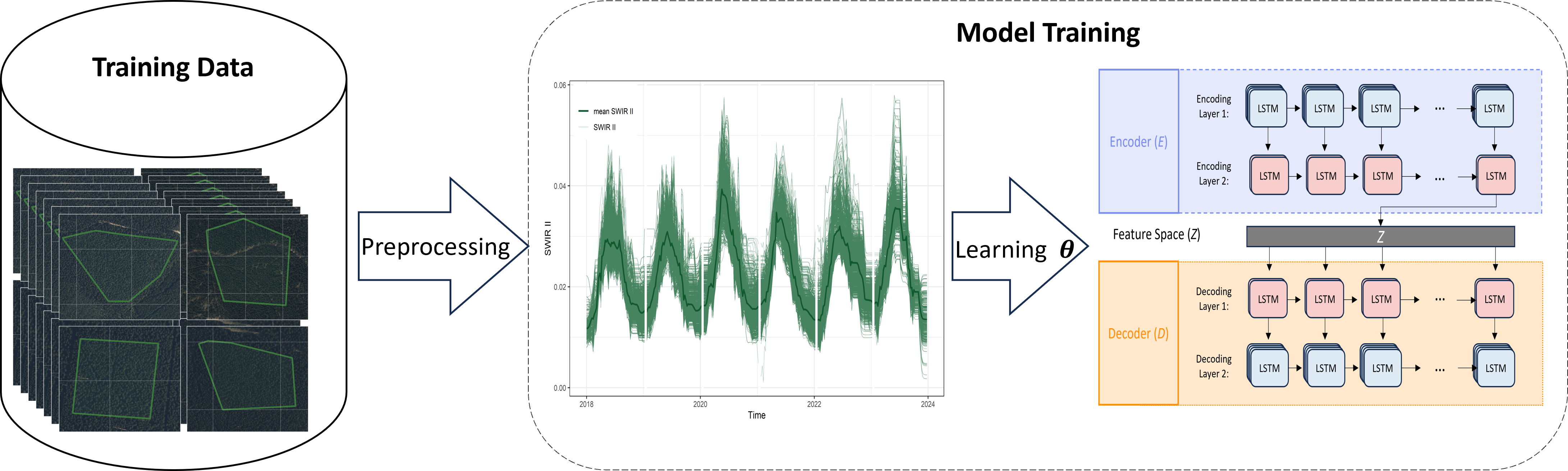}
    \caption{Process of Model Training}
    \label{training_model}
\end{figure}

All parameters are optimized during training to best reconstruct normal time series behavior \cite{goodfellow_deep_2016, wei_reconstruction-based_2023}. Fig. \ref{training_model} describes the operation in abstract terms.

The model was trained using the Keras API, specifically the compile and fit methods. The Adam optimization algorithm \cite{kingma_adam_2014} was employed for model updates, while mean squared error was used as loss function. The training process involves adjusting the model’s parameters - weights and biases - over multiple iterations to minimize reconstruction error.
Key training configurations included a batch size of 128, 200 epochs, and a validation split of 0.28. The batch size defined the number of training instances processed before updating parameters, while epochs determined the total number of dataset iterations.
\smallskip
This process was repeated for the window sizes of 26, 12 and 4 time steps, resulting in four models. 

\begin{figure}[H]
    \centering
    \includegraphics[width=0.9\linewidth]{{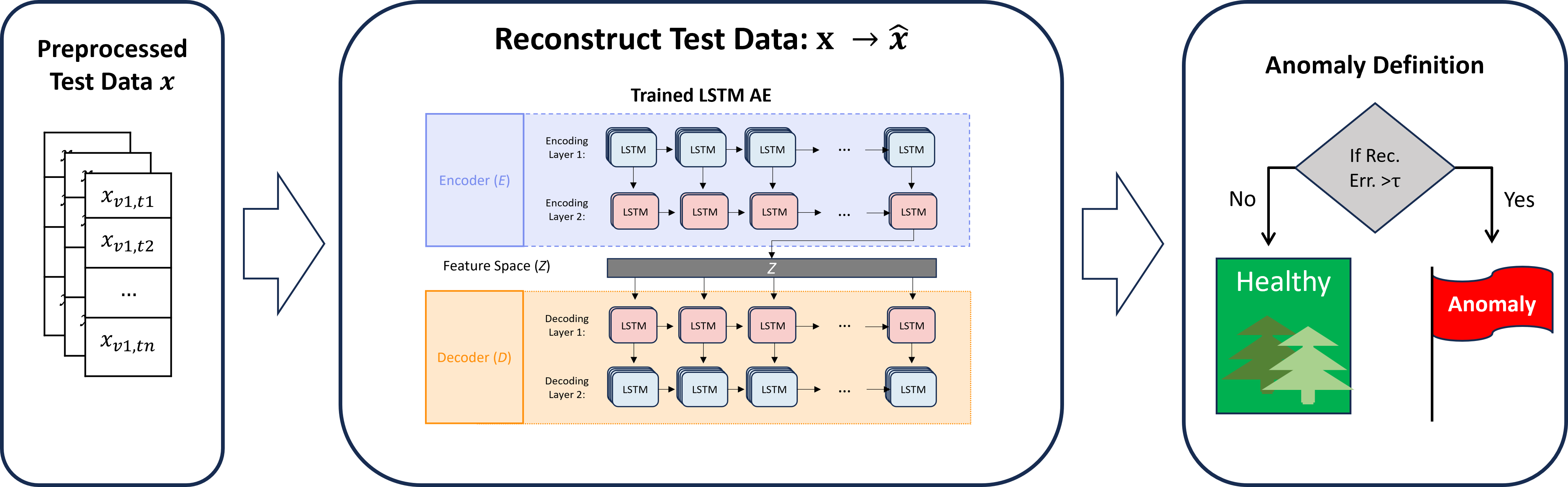}} 
    \caption{Process of Model Inference.}
    \label{testing_model}
\end{figure}

\paragraph{Model Inference}
During the inference phase, the models processes time sequences and compresses them into a lower-dimensional representation, known as the latent space. Afterward, the latent space is expanded to match the length of the input sequence, and this representation is then reconstructed by the decoder network \cite{goodfellow_deep_2016, russell_artificial_2020} (see Fig. \ref{testing_model}).

The data transformation within the \gls{ae} is regulated by the weights of the LSTM units, which are optimized during the training phase. Since these weights are trained to reconstruct healthy time series, the difference between the input $x$ and its reconstruction $\hat{x}$ will be larger if $x$ contains anomalous data. An anomaly at a time step $t$ is identified if the mean absolute error across all features, between $\hat{x}_t$ and $x_t$ exceeds a predefined threshold $\tau$ \cite{chandola_anomaly_2009}.

    \begin{align}
    Anomaly_t = \Bigg\{
        \begin{array}{ll}
        1, & \text{if } |x_t-\hat{x}_t| > \tau,\\
        0, & \text{otherwise.}
        \end{array}
\end{align}

\begin{figure}[H]
    \centering
    \includegraphics[width=0.9\linewidth]{{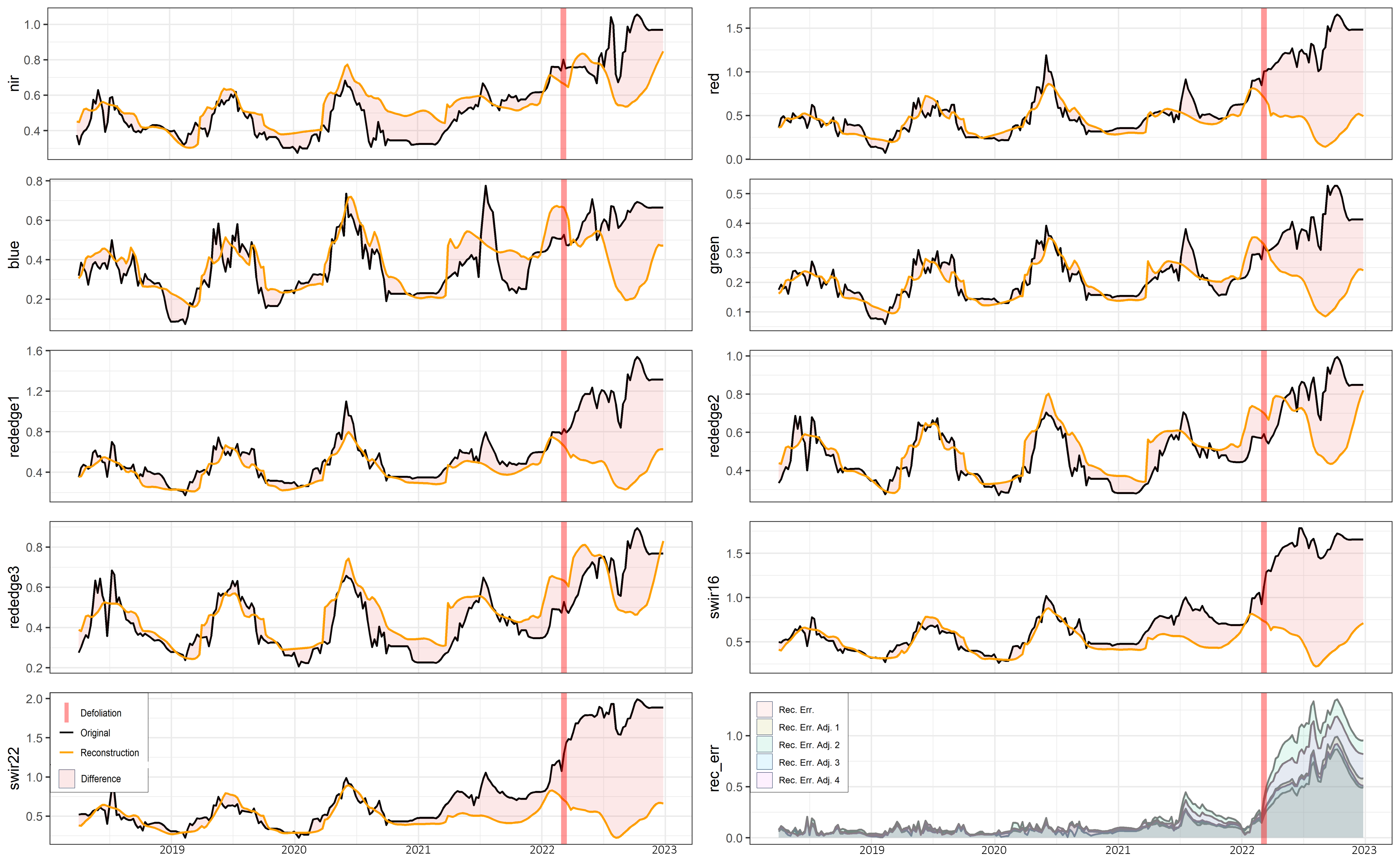}} 
    \caption{The original time series and its reconstructions across all features of a single time series, including reconstruction errors calculated using the MAE between different feature combinations.}
    \label{reconstruction}
\end{figure}

\medskip
For every model we reconstruct every multivariate time series within the test dataset. For every time series, each band's time series is reconstructed based on its own temporal pattern, the relationships between different bands, and their temporal dependencies.  Note that the model is able to reconstruct some bands better in healthy areas than in unhealthy forest areas, as shown in Fig. \ref{reconstruction}. 

Consequently, we calculated the adjusted reconstruction errors, which are based on the MAE of different band combinations. This adjustment is intended to help distinguishing between healthy and unhealthy time points. We selected the following combinations: 
\begin{itemize}
    \item \emph{rec\_err}: all bands
    \item \emph{rec\_err\_adj1}: NIR, RED, REDEDGE 1-3, SWIR I/II
    \item \emph{rec\_err\_adj2}: REDEDGE 1-3, SWIR II
    \item \emph{rec\_err\_adj3}: NIR, REDEDGE 1-3, SWIR I, SWIR II
    \item \emph{rec\_err\_adj4}: NIR, RED, SWIR I, SWIR II
\end{itemize}
\smallskip

\par
The predefined threshold was calculated individually for each window size by first applying the model to a set of time series that represent a healthy forest. The value for the 0.998 quantile of the reconstruction errors was then selected as the threshold. 

\subsection{Evaluation Framework}

As mentioned in section \ref{Related Work}, in remote sensing-based forest health monitoring, labeled data is extremely difficult to obtain. The exact start of the bark beetle infestation cannot be determined from remote sensing \cite{kautz_early_2024}. Therefore, evaluating models that detect early anomalies to predict forest disturbance is very difficult if there is no labeled data. 
We classified defoliation in pixels based on VI thresholds and created a scoring method that assumes an anomaly window around the defoliation date. The earlier anomalies are detected within the window, the higher they get rewarded. 
\medskip
\paragraph{Defoliation Threshold }
Several authors used static thresholds based on VIs to determine if spruce vegetation within a cell is defoliated \cite{otsu_estimating_2019, goldbergs_hierarchical_2024}. 
NDVI is frequently cited in the literature as a health indicator of forest vegetation \cite{ brovkina_unmanned_2018, goldbergs_hierarchical_2024, marx_sensitivity_2017,otsu_estimating_2019, rahimzadeh-bajgiran_detection_2018} and is an efficient measure to distinguish dead from healthy spruce trees \cite{brovkina_unmanned_2018,goldbergs_hierarchical_2024}. Although the 10-meter resolution of Sentinel-2 imagery does not allow the analysis of individual trees, changes in crown color and needle loss alter the NDVI value in a pixel. Previous studies suggest a NDVI range of 0.481-0.584 as threshold to assess defoliation in mixed conifer-deciduous stands\cite{otsu_estimating_2019}. More recent research by Goldbergs and Upenieks (2024) \cite{goldbergs_hierarchical_2024} evaluated an NDVI threshold of 0.46 to distinguish between living and dead trees within spruce monocultures. Through visual analysis of the images we found 0.53 to be a fitting threshold for the NDVI.

\begin{equation}
    NDVI = \frac{B8 - B4}{B8 - B4}
    \label{eq:ndvi}
\end{equation}
Additionally we selected the Tasseled Cap Wetness (TCW) index, which is a suiting index for disturbance detection on sentinel time series, especially being more precise during the winter months \cite{barta_early_2021}.
\begin{align}
    TCW &=  {\scriptstyle 0.1509} \cdot B2 + {\scriptstyle0.1973} \cdot B3 +{\scriptstyle 0.3279 }\cdot B4 \notag \\
    &\quad + {\scriptstyle0.3406 }\cdot B8 -{\scriptstyle 0.7112} \cdot B11 -{\scriptstyle 0.4572} \cdot B12 \label{eq:tcw}
\end{align}
We found a threshold of -0.03 to be especially suited. In our test data we searched for pixels that fall below the threshold for three consecutive weeks and classified the first instance as defoliation. A portion of the defoliation dates was visually checked and confirmed using Sentinel-2 images.  
\medskip
\paragraph{Anomaly Window}
Anticipating that the anomaly detection algorithm detects anomalies prior to the apparent defoliated state, we implemented a buffer zone to categorize infected cells before the needle loss or discoloration of the canopy (i.e. defoliation). We analyzed the lag between predicted anomaly and defoliation, to generate a buffer period preceding defoliation, during which anomalies are correctly labeled as such rather than erroneously classified as false positives. 
We chose a 52-week anomaly window to capture early indicators of forest health decline leading to defoliation. Bark-beetle infested trees show no evidence of differences in NDVI for up to 75 days after infestation \cite{kautz_patterns_2023, jamali_examining_2023}. Timely detection is further hampered by the delayed and irregular vitality response of spruce crowns as well as frequent cloud cover \cite{kautz_early_2024}, both of which we observed in our time series. 
Bark-beetles primarily attack trees that are already weakened by stress factors, such as reduced resin or sap pressure due to a disturbed water balance \cite{schwenke_beziehungen_1985}. In addition, our method detects general forest health anomalies - not just bark-beetle infestations - a full seasonal cycle allows us to capture phenological variations that signal tree weakening, improving the detection of true anomalies and enhancing the predictive ability of defoliation events. \par
If an anomaly is detected within the buffer interval around the defoliation, it is classified as a true positive (TP). If an anomaly is flagged outside the buffer, it is classified as a false positive (FP). If no anomaly is flagged within the buffer, it is classified as a false negative (FN). If a pixel's time series shows no defoliation and no anomaly is flagged, it is classified as a true negative (TN).
These results where utilized to calculate the accuracy, precision and F1 score, to evaluate the models performance: 
\begin{equation}
    \text{Accuracy} = \frac{\text{TP} + \text{TN}}{\text{TP} + \text{TN}+ \text{FP} + \text{FN}}
\end{equation} 
\begin{equation}   
    \text{Precision} = \frac{\text{TP}}
    {\text{TP} + \text{FP}} 
\end{equation} 
\begin{equation}   
     \text{Recall} = \frac{\text{TP}}{\text{TP} + \text{FN}} 
     \end{equation} 
\begin{equation}   
    \text{F1 Score} = 2 \times \frac{\text{Precision} \times \text{Recall}}{\text{Precision} + \text{Recall}} 
\end{equation} 
This methodical approach ensures a robust and transparent evaluation of the models, providing a solid foundation for determining the best-performing model\cite{powers_evaluation_2008}.

\begin{figure}[H]
    \centering
    \includegraphics[width=0.9\linewidth]{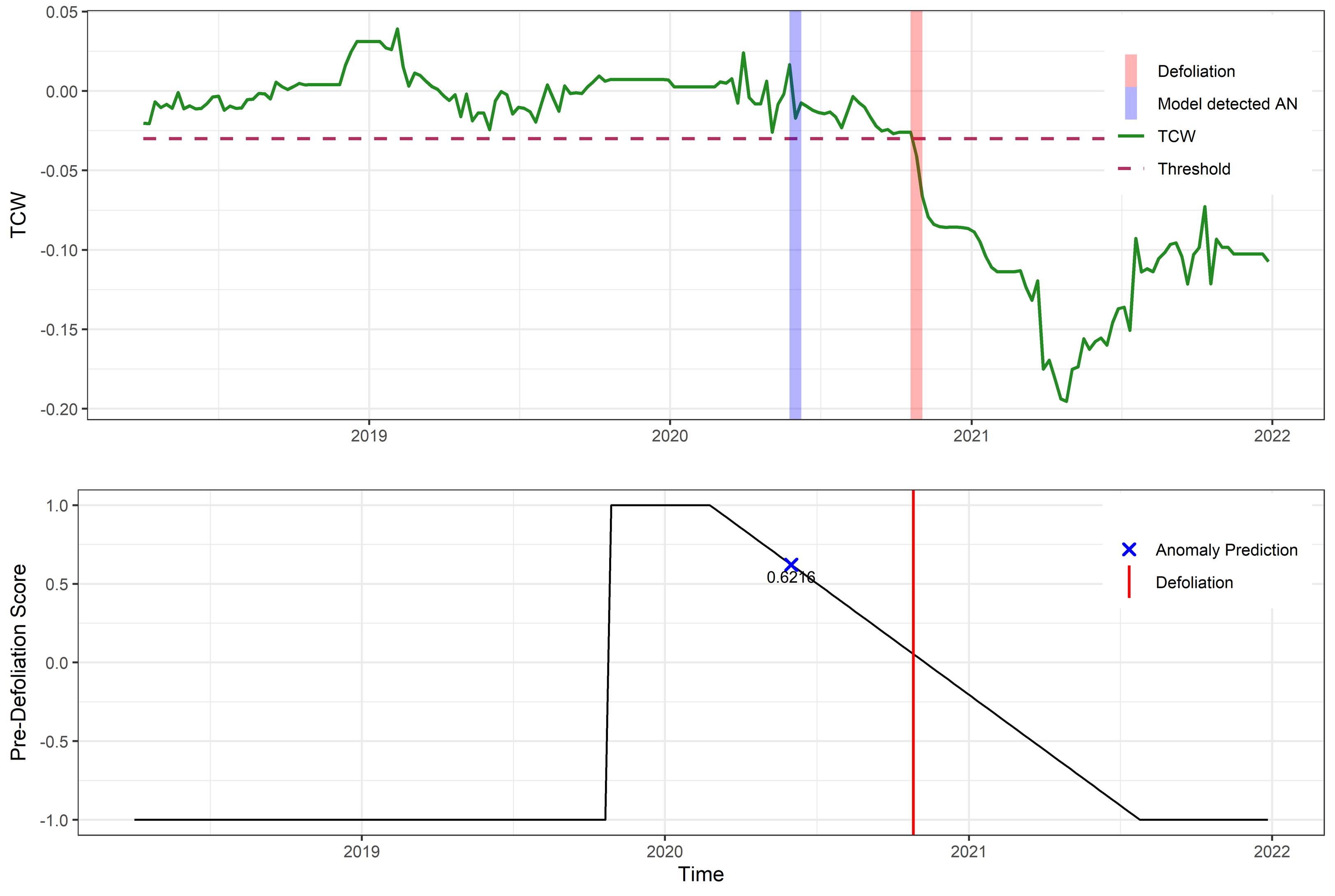}
    \caption{Pre-Defoliation Scoring system based on the temporal position of anomaly detection relative to defoliation. Early detections are rewarded, while late or false detections are penalized.}
    \label{pd_score_fig1}
\end{figure}

\medskip
\paragraph{Pre-Defoliation Score}
The above measures are commonly used in anomaly detection, but all fail to promote early detection, despite its critical importance in mitigating ecological damage. Kautz et al. (2024)\cite{kautz_early_2024} highlight that only 23\% of the studies on early bark-beetle detection rely on accurate ground truth data, underscoring the rarity of well-labeled datasets. We have developed a novel scoring method to address this gap: the \emph{Pre-Defoliation Score}. It is designed to evaluate and compare anomaly detection models in the context of forest disturbances when no labeled anomaly dates are available. This metric rewards early anomaly detection while penalizing both false positives and late detections, ensuring a more reliable assessment of model performance. See Fig. \ref{pd_score_fig1}.

To implement the Pre-Defoliation Score, we defined an anomaly window around the time of defoliation - i.e., the moment when vegetation loss is certain to have occurred. Anomalies detected earlier within this window receive a higher score, whereas detections following the defoliation event are assigned increasingly negative values. False detections, outside the anomaly window or in healthy time series, receive the lowest possible score (-1). Anomalies detected within the first third of the pre-defoliation window were assigned a score of 1, as we observed that detections in this time frame provide equally valuable early warnings. The score decreases linearly for later detections. It reaches zero two weeks after the defoliation event. This design accounts for practical constraints, such as cloud cover, which can hinder detection in earlier satellite images.  If an anomaly is detected beyond this time frame, it is assigned an increasingly negative score. Anomalies predicted outside the predefined anomaly window or in a healthy forest time series are penalized with a score of -1. Hence, if no anomaly is detected in a time series where no defoliation occurred (i.e. a healthy forest), the model is rewarded with a true negative (TN) score of 0.5. The Pre-Defoliation Score is defined by the following equation: 
\begin{equation}
\text{pd\_score($lag$)} = \frac{-3}{2\cdot k +s} \cdot \left( lag - s \right)
\label{eq:fx_continuous}
\end{equation}
where the difference between time of anomaly detection and defoliation time ($t_{\text{anomaly}} - t_{\text{defoliation}}$) is denoted as $lag$ and $k$ represents the anomaly window (52). A shift, denoted as $s$ is added to award an anomaly detected at the time of obvious defoliation with a positive value. 


\begin{align} 
    \text{pd\_score($lag$)} =\scriptstyle{ \begin{cases}-1, & \text{if }  |lag - s| > \frac{2}{3}k \\ 1, & \text{if } (lag - s) < -\frac{2}{3}k \land  lag >= -k \\ \text{pd\_score($lag$)}, & \text{otherwise} \end{cases}}
    \label{eq:fx_conditions}
\end{align}


If multiple anomalies were detected within the anomaly window, only the first anomaly was considered, while subsequent detections were assigned a score of zero to ensure that the models are not rewarded multiple times for a single defoliation event.

\medskip
The Pre-Defoliation Score provides a robust framework to evaluate early forest disturbance detection. By prioritizing timely anomaly identification while penalizing false alarms and delayed detections, this metric offers a meaningful way to compare models, even in the absence of perfectly labeled ground-truth data. Future studies can leverage this approach to improve remote sensing-based  forest health monitoring. 

\subsection{Model Scoring}

To address the research objective of analyzing the importance of temporal information, we tested LSTM-Autoencoder models using diverse input sequence lengths of 4, 12 , 24  and 52 weeks. The adjusted reconstruction errors of all four models were compared with each other.

To assess the performance advantages of our models compared to established methods, we compared the LSTM-AE models to the break detection algorithm BFAST. BFAST detects structural breaks in time series trends, making it particularly suitable for analyzing VIs. We tested the BFAST algorithm on the following VIs: $NDVI$, $DWSI$, $EVI$, and $DRS$, which are frequently used in bark-beetle detection \cite{verbesselt_detecting_2010,watts_effectiveness_2014, jamali_kernel-based_2024, huo_early_2021, trubin_detection_2024}.
BFAST requires a historical reference period for each observed pixel, we set the training phase from January 2018 to January 2020 to ensure a sufficient number of observations. The algorithm outputs a detected break time for each time series, which we then use to compute the Pre-Defoliation Score, along with the performance metrics precision, accuracy and F1-score. Finally, we compared these results to those of the LSTM-AE models to evaluate their relative effectiveness in early anomaly detection.

\section{Results and Discussion}

\subsection{Sequence Length and Adjusted Reconstruction Errors}
\begin{figure}[H]
    \centering
    \includegraphics[width=0.8\linewidth]{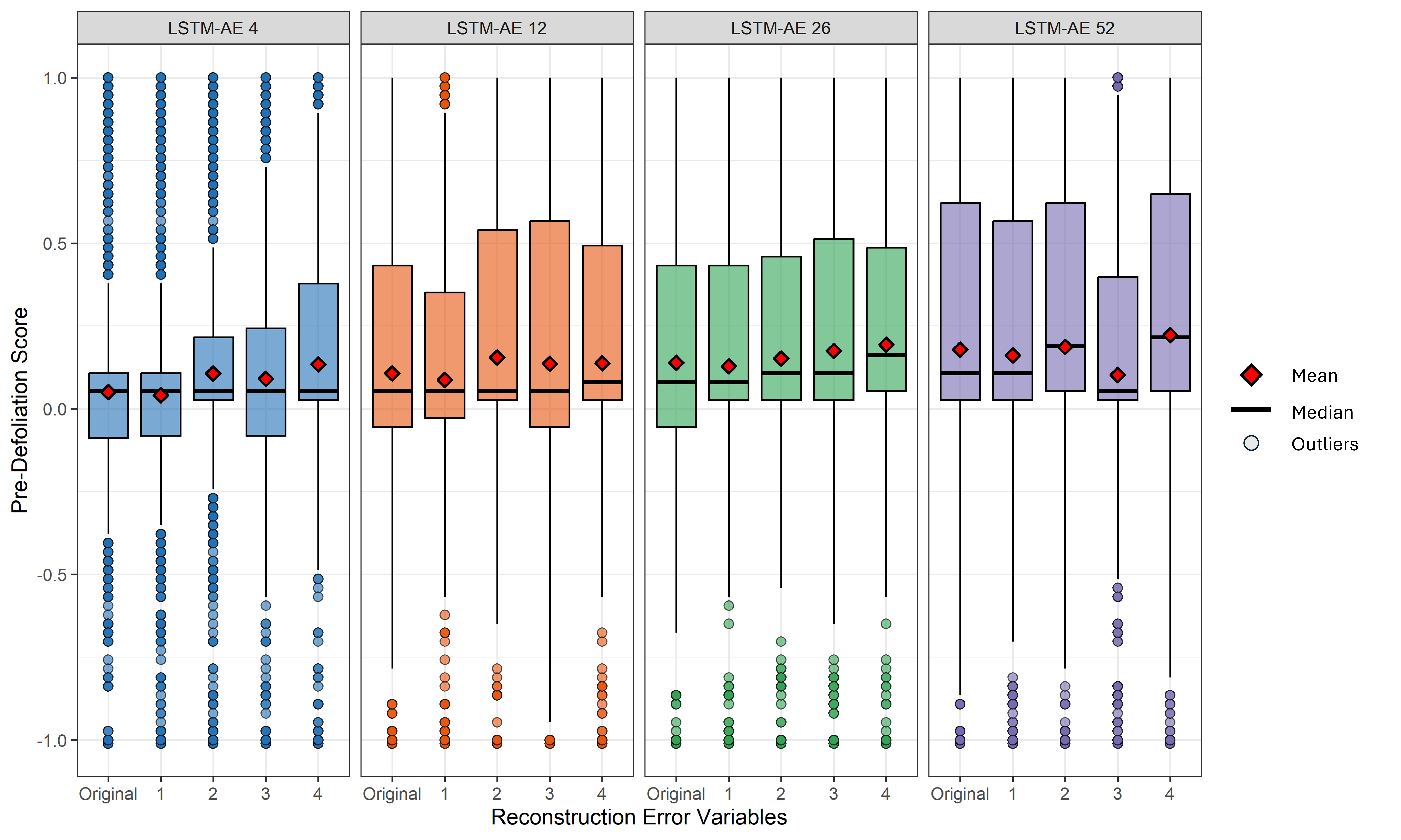} 
  \caption{ Pre-Defoliation Score comparison of LSTM model with different sequence lengths and different feature combinations.}
    \label{fig:bp1}
\end{figure}

The results in Fig. \ref{fig:bp1} show that \emph{rec\_err\_adj4} consistently achieved the highest Pre-Defoliation Scores across all LSTM-AE window sizes, indicating, that this combination of bands is most effective. Except for window size 12 it achieved the highest mean and median pd\_scores. Consequently, we used this feature combination for  further analyses. We excluded TNs from all box plots to prevent clustering at score 0.5.

\begin{figure}[h]
    \centering
    \includegraphics[width=0.8\linewidth]{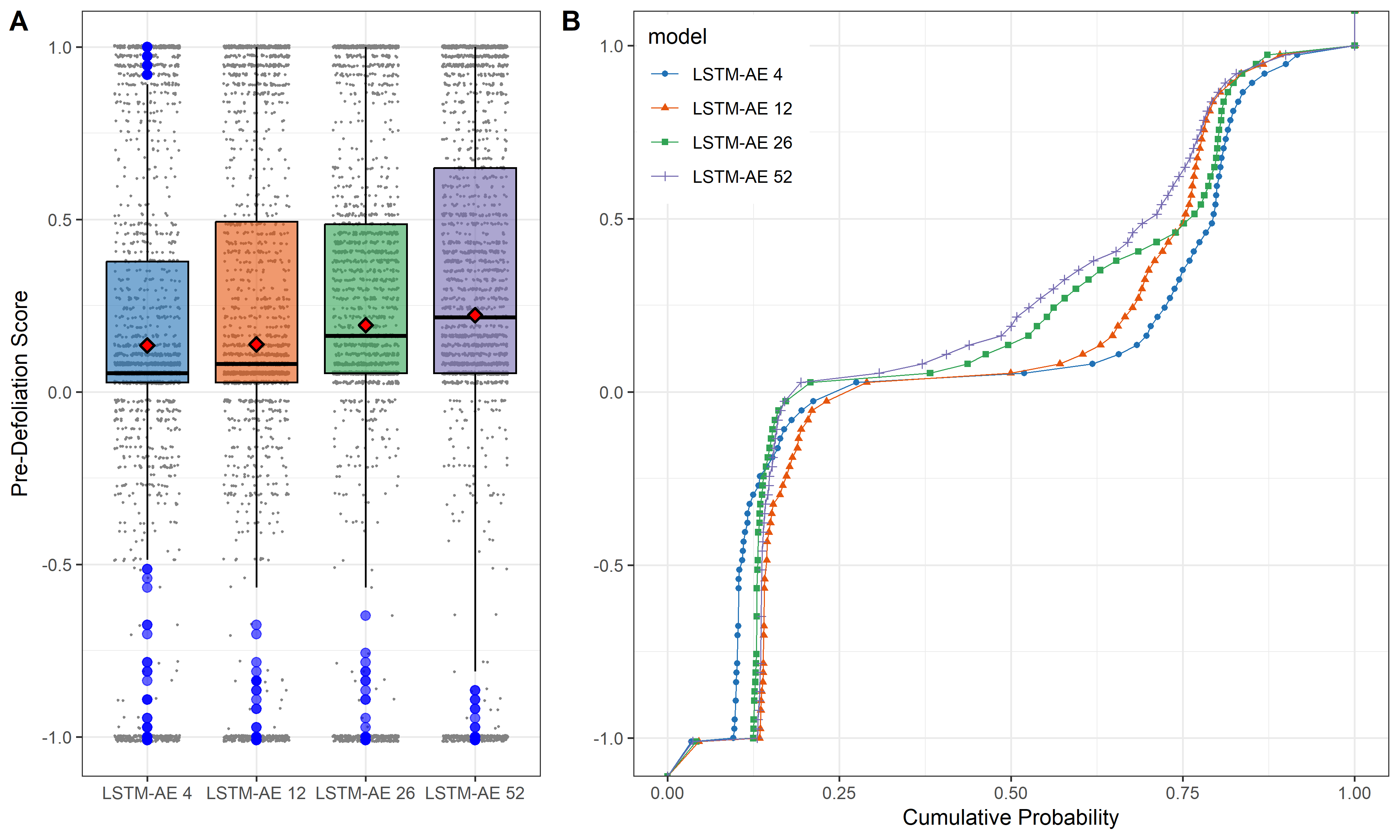} 
    \caption{Pre-Defoliation Score comparison using different sequence lengths in LSTM models. (A) Box plots illustrating the distribution of scores across different models. (B) Cumulative probability curves depicting the score distributions.}
    \label{bp_ecdf}
\end{figure}

We also found that window size had a notable effect on model performance. We will refer to these experiments as "LSTM-AE" concatenated with the respected window size. Models with longer sequence lengths (LSTM-AE 26 and LSTM-AE 52) exhibit higher median and mean pd\_scores, i.e.  a higher proportion of anomalies could be detected before defoliation. The LSTM-AE4 model has a smaller variance, fewer late detections, and false positives. However, it did not perform well in early detection. Fig. \ref{bp_ecdf} shows the distribution of pd\_scores more detailed and reveals which model flagged anomalies early or late.   

The pd\_score measures how early anomalies are detected relative to defoliation. A pd\_score of 0 means detection occurs at defoliation, while higher scores indicate earlier detection. Scores above 0.17 correspond to identification at least one month in advance.

Fig. \ref{bp_ecdf}a shows that larger window sizes generally yield higher pd\_scores, with the 52-week model demonstrating the most effective early detection, scoring the highest mean and median values. Fig. \ref{bp_ecdf}-B reinforces this trend, as models with smaller windows detect fewer anomalies in the 0–0.5 range.
Window size determines how temporal relationships between features are captured. Smaller windows (e.g., size 4) focus on short-term dependencies and relationships between feature. Larger windows incorporate broader patterns, creating a more abstract representation that captures temporal relationships more effectively. However, larger windows can also increase false positives, as reconstructions deviate further from the original input. Fluctuations in data—caused by factors like snow or cloud cover—may trigger false anomalies. Therefore, the optimal window size should be selected based on the specific task.

\subsection{Comparison to BFAST}

To compare the performance of our LSTM-AE models and the BFAST algorithm across different VIs, we calculated the mean Pre-Defoliation Score and determined the percentage of pixels where anomalies were flagged more than four weeks prior to defoliation (i.e. score $>$ 0.17). Additionally, we computed accuracy, precision, and F1-score to further assess model performance. Because BFAST uses two years as training period, we excluded pixel time series, where defoliation occurred before 2020 from the evaluation. The results are presented in Table \ref{tab_perf}.

\begin{table}[H]
\caption{Performance Comparison of LSTM-AE and BFAST Models\label{tab_perf}}
\centering
\resizebox{\linewidth}{!}{  
\begin{tabular}{|c|cccc|cccc|}
\hline
\textbf{Metric} & \multicolumn{4}{c|}{\textbf{LSTM-AE}} & \multicolumn{4}{c|}{\textbf{BFAST}} \\
& \textbf{4} & \textbf{12} & \textbf{26} & \textbf{52} & \textbf{DRS} & \textbf{DWSI} & \textbf{EVI2} & \textbf{NDVI} \\
\hline
\textbf{Precision} & \textbf{0.884} & 0.842 & 0.861 & 0.853 & 0.808 & 0.662 & 0.512 & 0.722 \\
\textbf{Accuracy} & 0.837 & 0.827 & \textbf{0.872} & 0.871 & 0.716 & 0.630 & 0.435 & 0.593 \\
\textbf{F1-Score} & 0.876 & 0.866 & \textbf{0.903} & 0.901 & 0.771 & 0.692 & 0.392 & 0.620 \\
\textbf{Mean PDS} & 0.225 & 0.229 & 0.272 & \textbf{0.297} & 0.216 & 0.078 & -0.150 & 0.108 \\
\textbf{PDS \%} & 33.54 & 41.73 & 56.74 & \textbf{61.12} & 45.27 & 39.09 & 19.23 & 35.24 \\
\hline
\end{tabular}
}  
\end{table}

Table \ref{tab_perf} shows, that LSTM-AE4 delivers the highest precision because it is less prone to false positive (FP) detections. LSTM-AE26 achieves the highest accuracy and F1-Score among all models, with a slight advantage over LSTM-AE52. LSTM-AE52, it performs best in terms of mean Pre-Defoliation Score and detects the highest percentage of TP's over the four weeks preceding defoliation.
The BFAST detection method performs significantly worse; however, the DRS index yields the best results among the BFAST models.

\begin{figure}[h]
    \centering
    \includegraphics[width=0.8\linewidth]{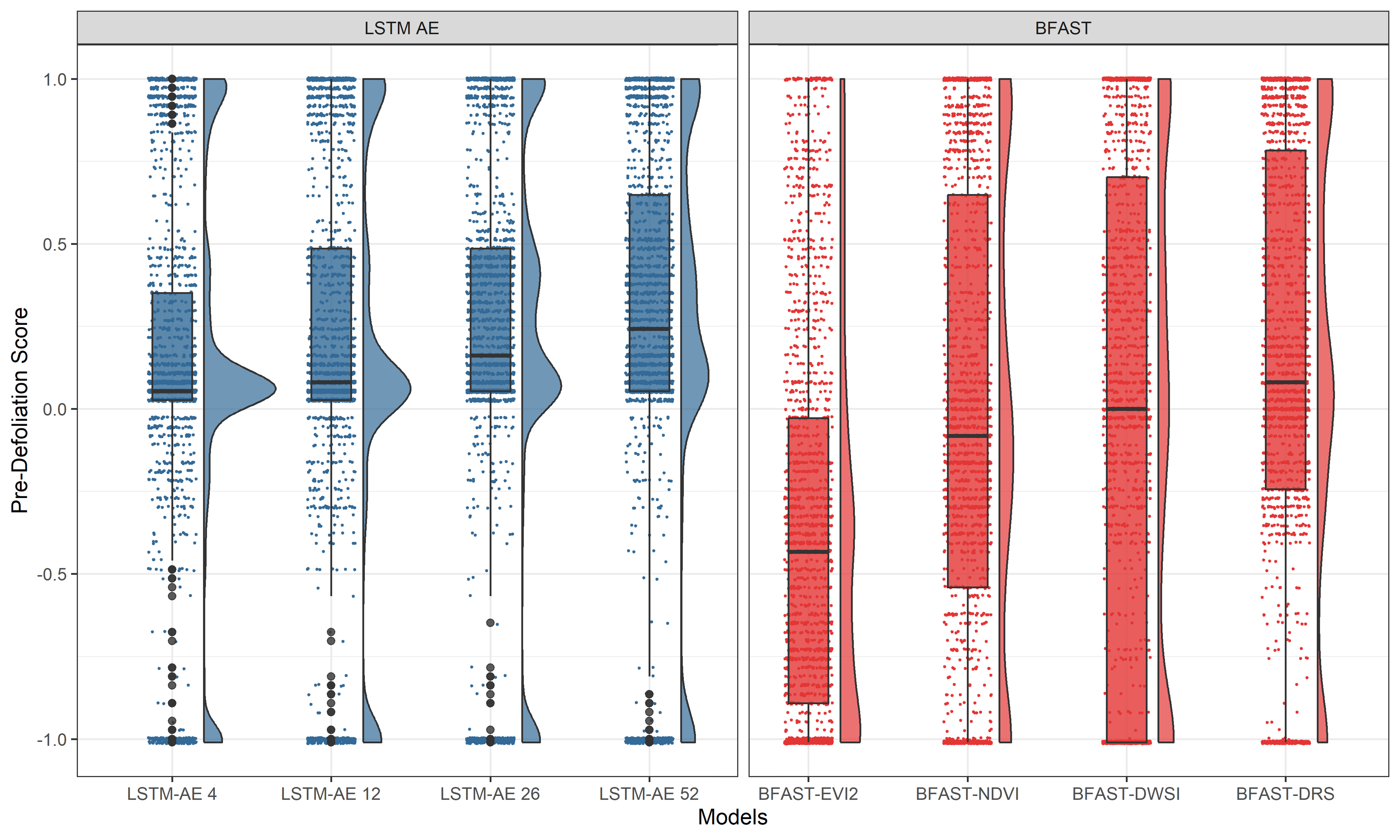}
    \caption{Pre-Defoliation Score Distribution for LSTM-AE and BFAST Models.}
    \label{bp_bf_lstmaae}
\end{figure}

The distribution of Pre-Defoliation Scores across our LSTM-AE models and BFAST on various VIs is shown in Fig. \ref{bp_bf_lstmaae}.

Fig. \ref{bp_bf_lstmaae} visualizes the distribution of scores achieved on the test dataset for each model. The blue LSTM-AE models generally exhibit lower variance, with scores clustering slightly above a pd\_score of 0. As window sizes increase, a greater number of higher scores are achieved, indicating that more anomalies are detected well before defoliation. Across the LSTM-AE models, there are accumulations of pd\_scores at 1 and smaller accumulations at -1. The clusters at 1 correspond to anomalies detected between 35 and 52 weeks before defoliation, all receiving the maximum score. Anomalies detected even earlier were assigned a score of -1, as they were considered false positives.

In contrast, the distribution of scores across the BFAST models (red) exhibits higher variance, reflecting greater variability in detection performance. While only a small percentage of detections in the LSTM-AE models are classified as false positives (pd\_score of -1), the BFAST models show a higher accumulation in this range. This trend is illustrated in Fig. \ref{bp_bf_lstmaae}, where a pd\_score of -1 appears as an outlier in the box plots for the LSTM Autoencoder models but falls within the whiskers for BFAST, indicating a more frequent occurrence of false positives.
Among the BFAST models, the $DRS$ VI delivers the best results. BFAST-DRS has a positive median, therefore over 50\% of detections were before the defoliation. Furthermore, it exhibits a high density in the upper score regions, greater than 0.5. This demonstrates its capability to detect changes at early stages.

\subsection{Evaluation Maps}

\begin{figure}[H]
    \centering
    \includegraphics[width=1\linewidth]{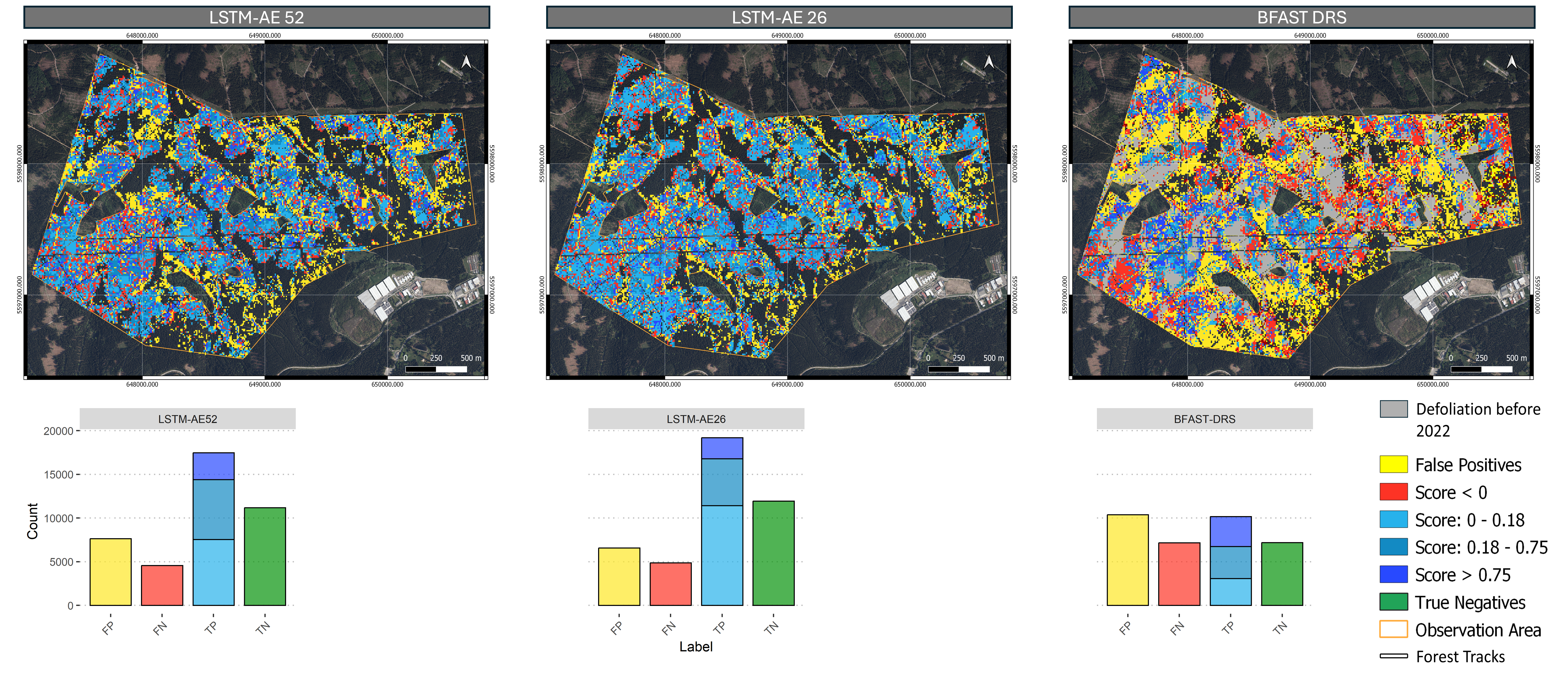}
    \caption{Pre-defoliation score maps of a Spruce cultivation in Thuringian Forest.}
    \label{score_maps}

    \centering
    \includegraphics[width=0.6\linewidth]{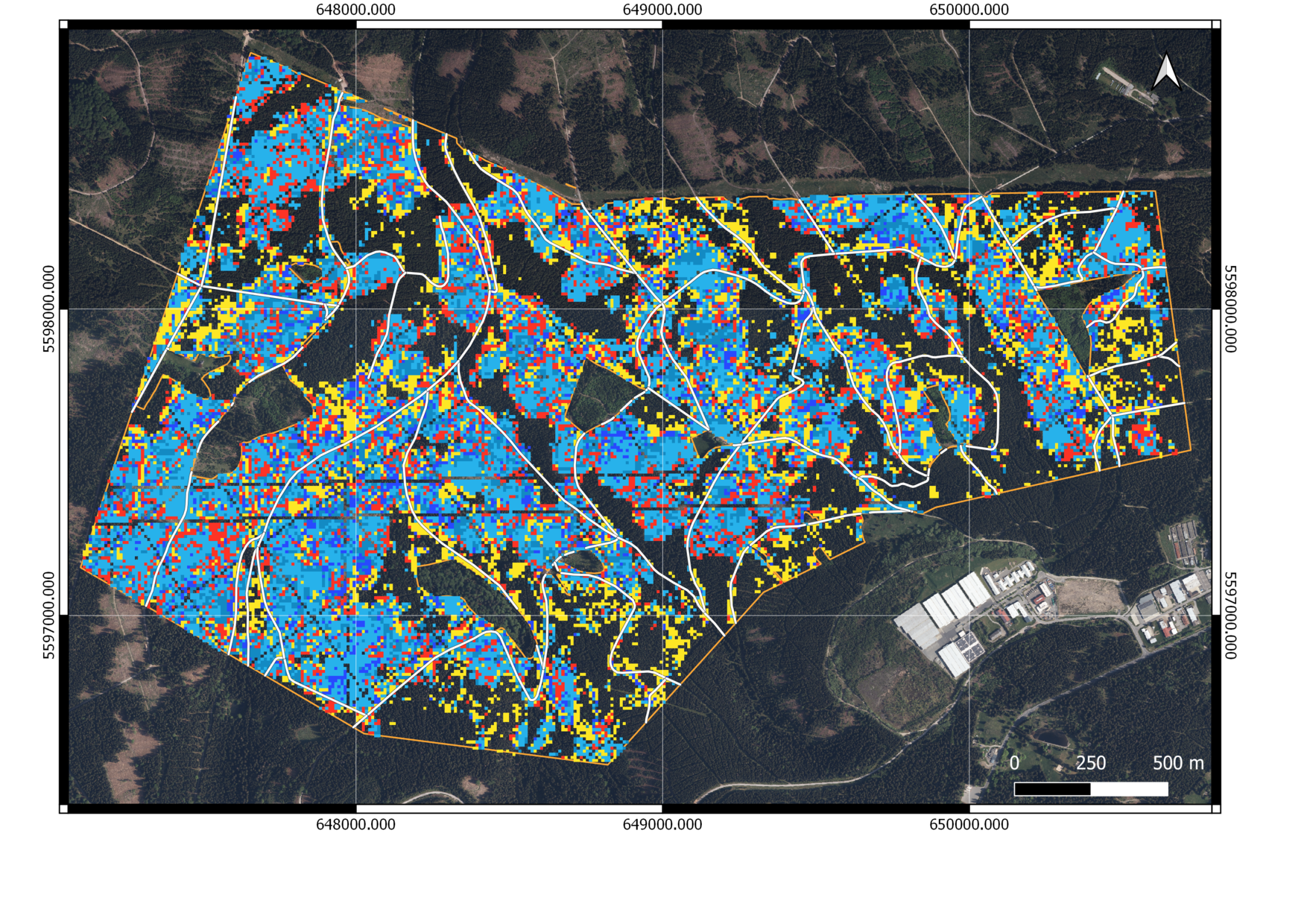}
    \caption{LSTM-AE 26 Score Map Forest Tracks}
    \label{map_foresttracks}
\end{figure}

The Pre-Defoliation Scoring system enables the creation of maps that visualize model performance within an observed area. Fig. \ref{score_maps} visualizes the Pre-Defoliation Scores of the respective pixels in a larger spruce cultivation area for the time span from 2020 to the end of 2024. 

The LSTM-AE models clearly demonstrate superior performance compared to the BFAST algorithm. LSTM-AE26 exhibits notably more robust anomaly detection and produces fewer false positives (yellow pixels) or late detections (red Pixels) than LSTM-AE52. However, LSTM-AE52 performs best in terms of detections during the four weeks preceding defoliation (pd\_score > than 0.17).
Many false positives (FPs) in the LSTM-AE models match forest tracks (see Fig. \ref{map_foresttracks}). This means, the model flags an anomaly because these pixels do not represent actual forest. For further research, forest roads need to be excluded from the data.

\begin{figure}[H]
\centering
\includegraphics[width=1\linewidth]{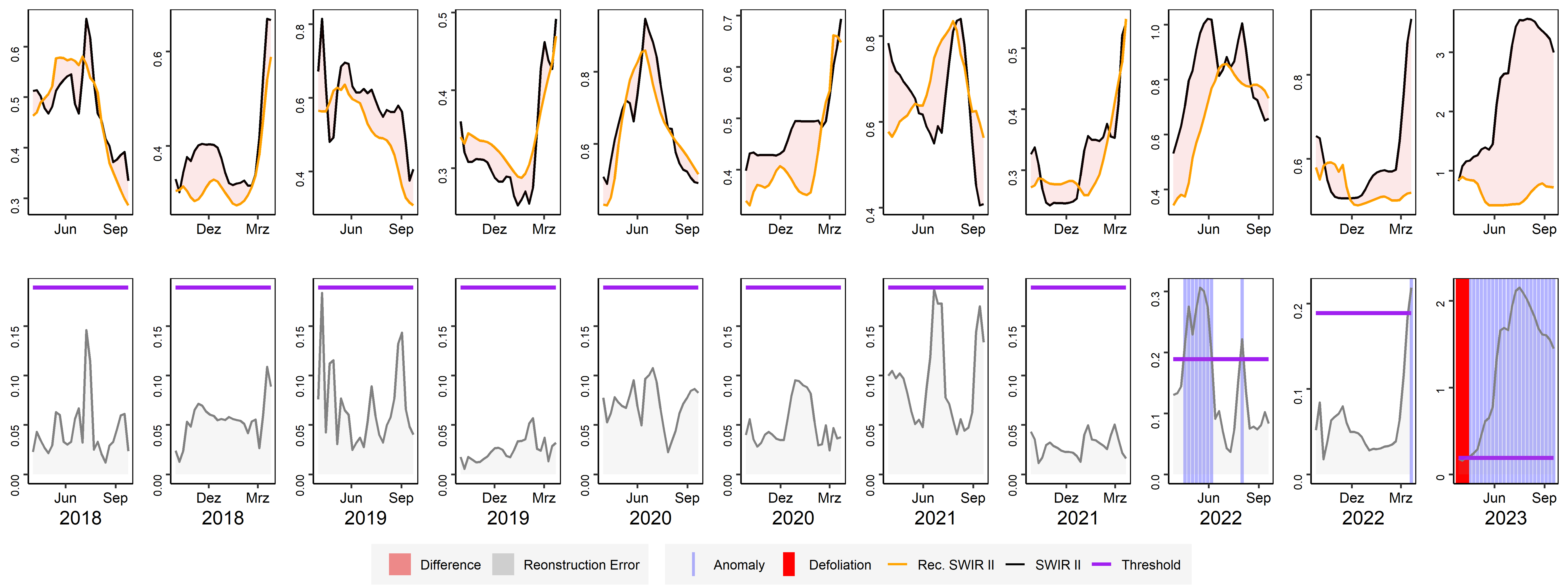}
\caption{Anomaly detection in short time series using LSTM-AE26.}
\label{ts_sequenc_lstm26}
\end{figure}

LSTM-AE52 outperforms LSTM-AE26 in very early detection. However, LSTM-AE26 is more robust against false detections and still provides strong results in anomaly detection. Hence, if the primary goal is to detect anomalies as early as possible while allowing only occasional false positives, we recommend using a window size of 26. This approach reduces the need for extensive input data, while ensuring an efficient analysis on the temporal context and maintaining robustness. It can reliably detect anomalies early with only 26 consecutive imputed time steps. Figure \ref{ts_sequenc_lstm26} illustrates the ability of LSTM-AE26 to detect anomalies before defoliation, on short input time series.

The upper row displays the original data and its reconstruction for a single feature, while the bottom row shows the corresponding reconstruction error across the rec\_err\_adj4 features in the observed time series sequence. LSTM-AE26 is applied individually to each window. By analyzing the temporal relationships between all variables, the model successfully detects anomalies prior to defoliation.
\smallskip

We employed the LSTM-AE26 model to generate an anomaly detection map, which provides a spatial and temporal representation of detected anomalies within a given area. This visualization allows for a more intuitive interpretation of when and where disturbances occur, supporting in the assessment of forest health over time.

\begin{figure}[H]
    \centering
\includegraphics[width=0.7\linewidth]{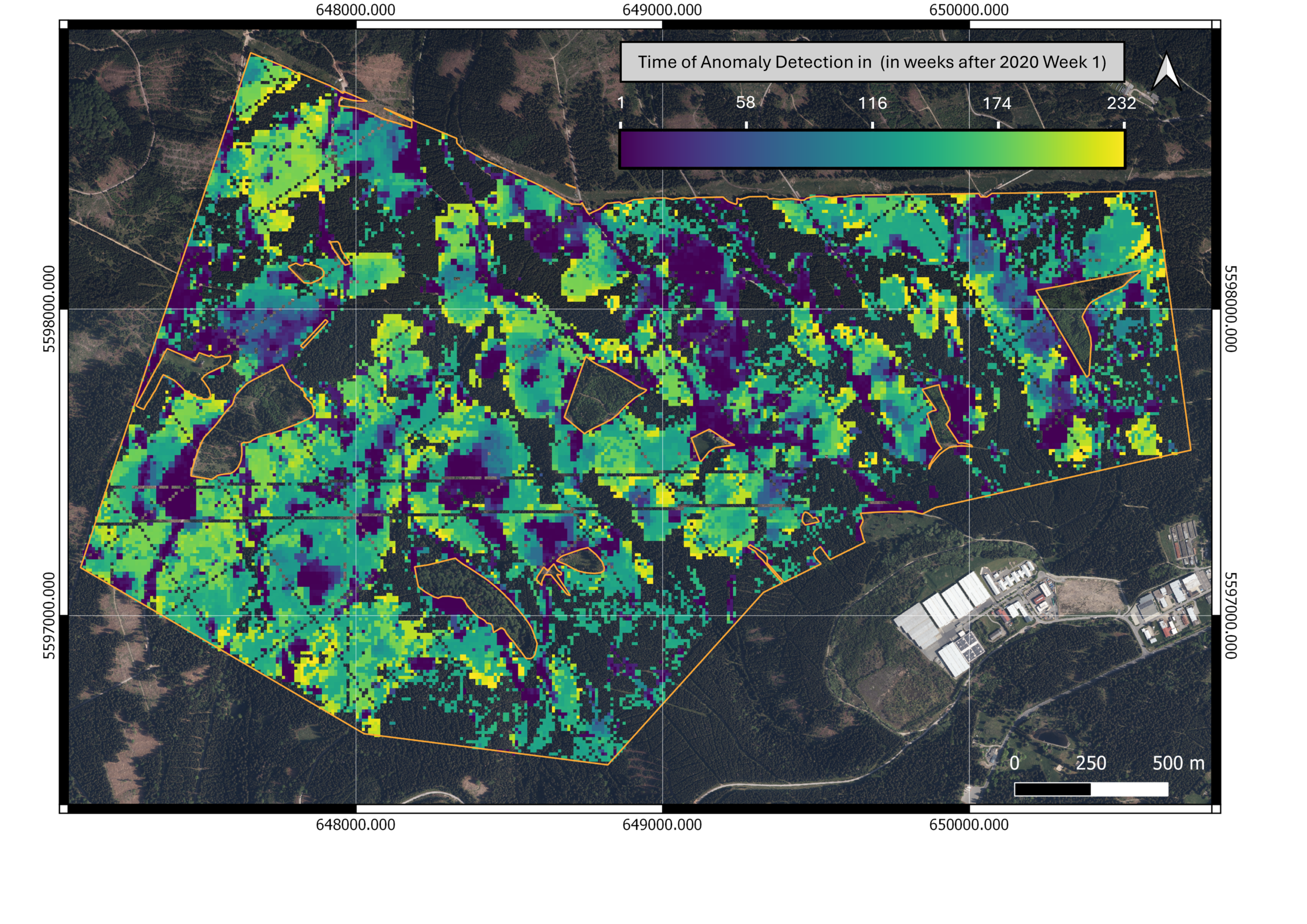}
    \caption{Anomaly Detection Map}
    \label{and_map}
\end{figure}
Fig. \ref{and_map} represents an anomaly map, where the gradient change symbolizes the weeks after the beginning of the observation, in which anomalies are detected. Applying these models can provide an understanding of disturbance patterns and enable an early intervention of secondary pest outbreaks through active forest management.

\section{Conclusion}
We introduced an anomaly detection framework for monitoring forest health dynamics in spruce cultivations, using bark-beetle outbreaks as a case study. In our work  we use an unsupervised Deep Learning algorithm, more specifically a multilayer LSTM Autoencoder trained on Sentinel-2-based time series. The model was tested with different input window sizes, revealing that a larger window size leads to earlier detections. Notably, the LSTM-AE52 variant achieved an accuracy of 87\%, with 61\% of the detected anomalies flagged more than four weeks before visible defoliation occurred in the pixel. Additionally, we found that a window size of 26 time steps was sufficient to ensure robust and early anomaly detection. The memory-efficient design of our model eliminates the need for labeled training data or extensive historical records for each pixel, making it well-suited for large-scale monitoring. By capturing both spectral and temporal relationships across all Sentinel-2 bands, our method provides a powerful tool for detecting subtle changes in forest health, that can forcast defoliation events. The proposed models were evaluated against BFAST and the results of this comparison highlight the superiority of the LSTM Autoencoder approach, particularly in terms of early anomaly detection. Furthermore, we introduced a novel scoring method that complements conventional performance metrics by rewarding early detections, a crucial factor in disturbance prediction tasks. This work serves as a foundation for future research, enabling the exploration of different forest structures and species, as well as further deep learning applications in SITS with similar input window sizes. Furthermore, our findings lay the groundwork for the implementation of LSTM Autoencoders as a continuous early warning system for forest health monitoring, with potential extensions to other environmental anomaly detection tasks.

\bibliographystyle{unsrt}  

\bibliography{filtered}





\end{document}